\documentclass[sigconf,9pt]{acmart}

\usepackage{xspace}
\usepackage{subfigure}
\usepackage{graphicx}
\usepackage{multirow}
\usepackage{pifont}

\AtBeginDocument{%
  }

\renewcommand\footnotetextcopyrightpermission[1]{}
\ifodd 1
\newcommand{\com}[1]{\textbf{\color{red}(COMMENT: #1)}} %comment of the 
\else

\newcommand{\com}[1]{}
\fi

\def\fig{Fig.}
\def\sec{Section }
\def\sysname{\textsf{EventPro}\xspace}
\def\tab{Table}
\def\eg{e.g.}
\def\ie{i.e.}

\begin{document}

%%
%% The "title" command has an optional parameter,
%% allowing the author to define a "short title" to be used in page headers.
\title{Count Every Rotation and Every Rotation Counts: Exploring Drone Dynamics via Propeller Sensing}

%%
%% The "author" command and its associated commands are used to define
%% the authors and their affiliations.
%% Of note is the shared affiliation of the first two authors, and the
%% "authornote" and "authornotemark" commands
%% used to denote shared contribution to the research.
% \author{Ben Trovato}
% \authornote{Both authors contributed equally to this research.}
% \email{trovato@corporation.com}
% \orcid{1234-5678-9012}
% \author{G.K.M. Tobin}
% \authornotemark[1]
% \email{webmaster@marysville-ohio.com}
% \affiliation{%
%   \institution{Institute for Clarity in Documentation}
%   \city{Dublin}
%   \state{Ohio}
%   \country{USA}
% }

\author{
Xuecheng Chen$^{1*}$, 
Jingao Xu$^{2*}$, 
Wenhua Ding$^1$, 
Haoyang Wang$^1$, 
Xinyu Luo$^1$,
Ruiyang Duan$^3$, 
Jialong Chen$^3$, 
Xueqian Wang$^1$, 
Yunhao Liu$^4$, 
Xinlei Chen$^{1,\dag}$
}

\affiliation{
\institution{
$^1$Shenzhen International Graduate School, Tsinghua University, China;
$^2$The University of Hong Kong;\\
$^3$Meituan Academy of Robotics Shenzhen, China;
$^4$School of Software, Tsinghua University, China;}
\country{}
\city{}
% Email: \{chenxc24, dingwh24, haoyang-22, luo-xy23, yunhao\}@mails.tsinghua.edu.cn, \\
% jingaox@andrew.cmu.edu, \{chenjialong02, duanruiyang\}@meituan.com, \{wang.xq, chen.xinlei\}@sz.tsinghua.edu.cn
}

\email{{chenxc24, dingwh24, haoyang-22, luo-xy23, yunhao}@mails.tsinghua.edu.cn}
\email{jingaox@andrew.cmu.edu, {chenjialong02, duanruiyang}@meituan.com, {wang.xq, chen.xinlei}@sz.tsinghua.edu.cn}

\thanks{$^*$ Co-primary authors.}
\thanks{$\dag$ Xinlei Chen is the corresponding author.}

%%
%% By default, the full list of authors will be used in the page
%% headers. Often, this list is too long, and will overlap
%% other information printed in the page headers. This command allows
%% the author to define a more concise list
%% of authors' names for this purpose.
\renewcommand{\shortauthors}{Xuecheng Chen et al.}

%%
%% The abstract is a short summary of the work to be presented in the
%% article.

\begin{abstract}
As drone-based applications proliferate, paramount contactless sensing of airborne drones from the ground becomes indispensable.
This work demonstrates concentrating on propeller rotational speed will substantially improve drone sensing performance and proposes an event-camera-based solution, \sysname.
\sysname features two components:
\textit{Count Every Rotation} achieves accurate, real-time propeller speed estimation by mitigating ultra-high sensitivity of event cameras to environmental noise.
\textit{Every Rotation Counts} leverages these speeds to infer both internal and external drone dynamics.
Extensive evaluations in real-world drone delivery scenarios show that \sysname achieves a sensing latency of 3$ms$ and a rotational speed estimation error of merely 0.23\%.
Additionally, \sysname infers drone flight commands with 96.5\% precision and improves drone tracking accuracy by over 22\% when combined with other sensing modalities.
\textit{
Demo: {\color{blue}\url{https://eventpro25.github.io/EventPro/}.}
}
\end{abstract}

\maketitle

% \input{intro}
% \vspace{-0.1cm}
\section{Introduction}\label{sec:intro}
\begin{figure}[t]
    \setlength{\abovecaptionskip}{0.2cm} % height above Figure X caption
    \setlength{\belowcaptionskip}{-0.3cm}
    \centering
    \includegraphics[width=0.93\columnwidth]{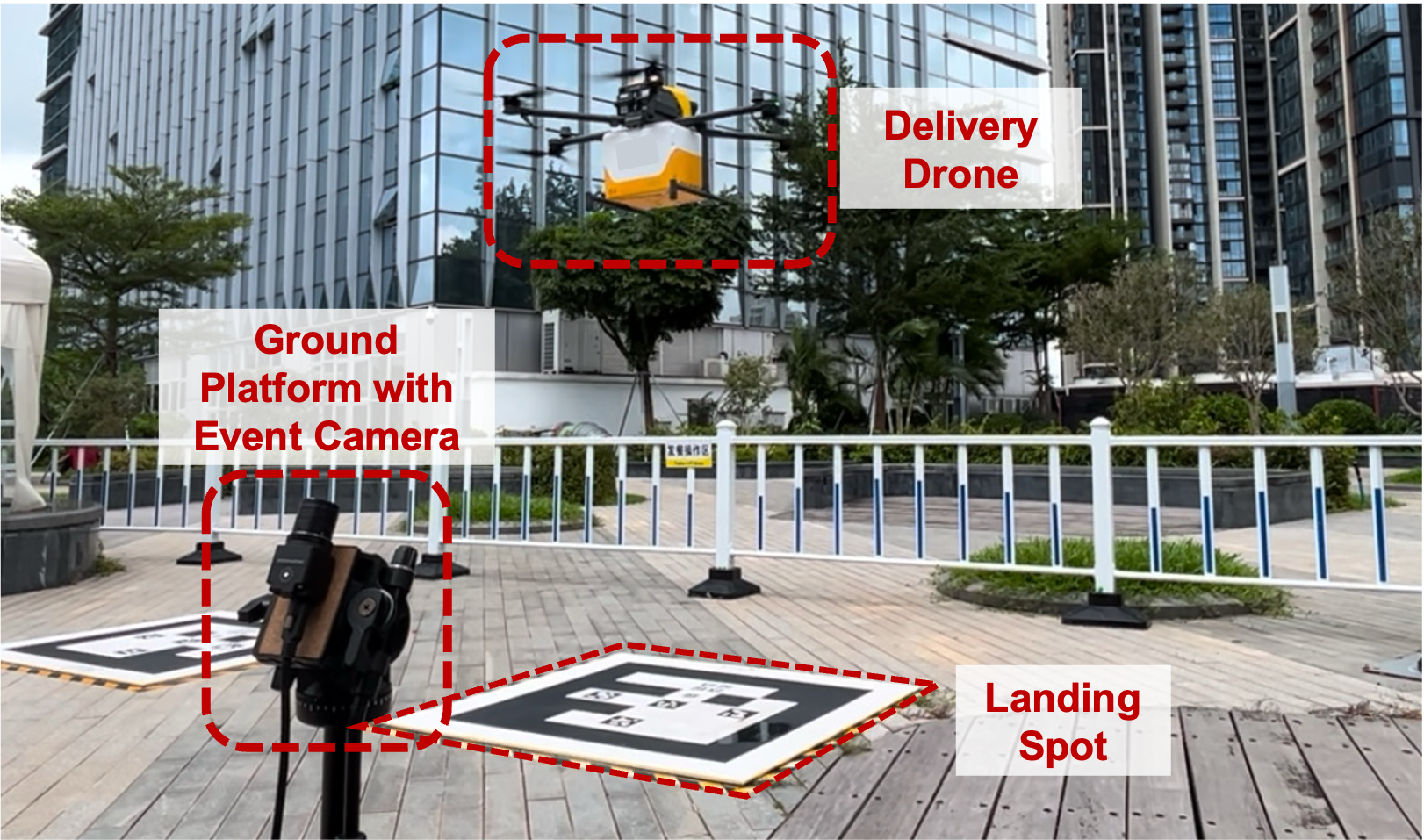}
    % \caption{\sysname locates a delivery drone and guides it to land on the landing spot. }
    \caption{\sysname locates a real-world delivery drone and guides it to land on the airport landing spot. }
    \label{fig:fig1}
    \vspace{-4mm}
\end{figure}

The drone-driven low-altitude economy, anticipated to reach a \$1 trillion valuation by 2040~\cite{chinadaily, zhang2025citynavagent, chen2024ddl}, is reshaping industries through transformative services like instant delivery~\cite{chen2024ddl,sharp2022authentication,zhao2024foes}, precision industrial monitoring~\cite{ben2020edge,xia2023anemoi,ahmad2023aerotraj}, and emergency response operations~\cite{schedl2021autonomous, zhang2023rf,sie2023batmobility}. 
Central to this rapidly expanding industry is the critical \textit{landing phase} (\fig\ref{fig:fig1}), during which ground platforms must rapidly identify drones descending from below 50 meters and guide them to precise touchdown at designated locations~\cite{he2023acoustic, sun2022aim}.

Current contactless, ground-based localization systems for drone landing primarily install cameras~\cite{zhu2018visdrone,petrides2017towards}, acoustic sensors~\cite{bannis2020bleep, wang2022micnest, bai2022spidr}, and radars~\cite{drozdowicz201635, shin2016distributed, hoffmann2016micro} at the center or along the edges of drone landing pads. These methods excel in low-speed, stable flights but face challenges in complex scenarios involving dynamic accelerations. Such performance degradation stems from the sensors' limited spatio-temporal resolution~\cite{wang2025mme}. For instance, radar often perceives drones merely as point objects and fails to accurately capture 6-degree-of-freedom rotational movements critical for precise landing adjustments \cite{herschfelt2017consumer}. Similarly, cameras operating at 20-60 FPS frequently may have difficulty profiling high-speed actions that occur during frame intervals~\cite{gehrig2024low, jiang2021flexible, jain2019scaling}.

\begin{figure}[t]
    \setlength{\abovecaptionskip}{0.15cm} % height above Figure X caption
    \centering
    \includegraphics[width=0.93\columnwidth]{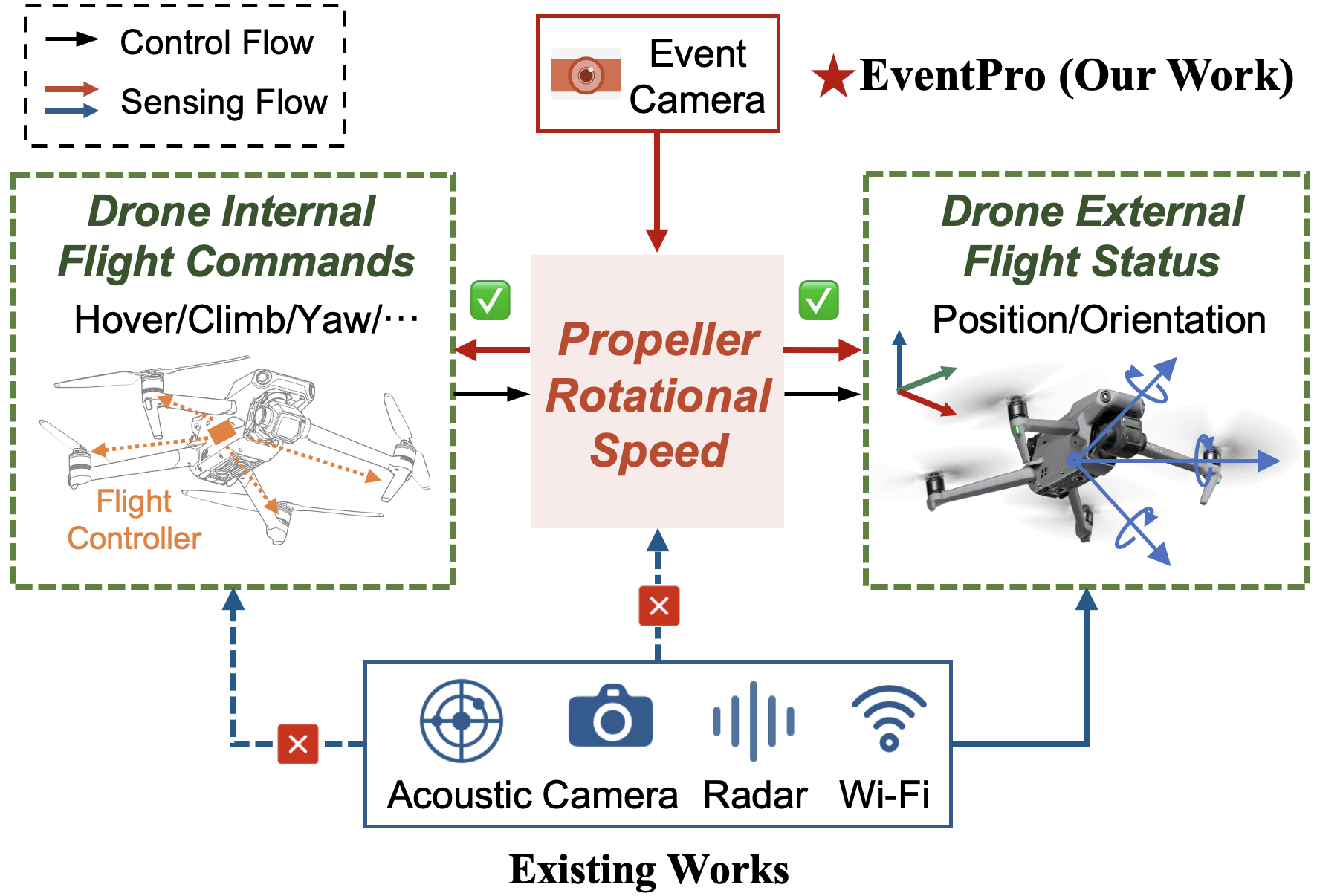}
    \caption{
    Propeller rotational speed bridges internal flight commands and external flight status.
    % Rethinking the drone status sensing problem from the entire flight control pipeline view.
    % \textnormal{Our work highlights propeller rotational speeds as the critical link between internal flight commands and external flight status. 
    % Our proposed system, \sysname, capitalizes on this connection to enhance bidirectional sensing capabilities, improving the sensing ability of both internal and external drone dynamics compared to existing works.
    % }
    }
    \label{fig:fig2}
    \vspace{-5mm}
    % \vspace{-6.09mm}
\end{figure}

To overcome these limitations, we expand our perspective beyond individual sensors or algorithms to encompass the full drone propulsion and flight control system, as shown in \fig\ref{fig:fig2}.
Given a user-desired or planner-calculated flight trajectory (\ie, a continuous location, linear/angular velocity, and acceleration over time), the drone’s position and attitude controllers produce \textit{internal flight commands}\footnote{Commercial flight controllers such as PX4~\cite{px4} standardize drone flight commands in formats like \textit{hovering}, \textit{climbing/descending} a specific number of meters, and adjusting \textit{pitch/yaw/roll} by predetermined degrees~\cite{mobilesdk}.} 
at every instant~\cite{mahony2012multirotor}. These commands guide each propeller’s motor controller to adjust blade speeds, altering forces and torques, eventually adjusting the drone's \textit{external flight status} (\ie, velocity, position, and orientation).

In short, the drone's \textit{external flight status} is governed by its \textit{internal flight commands}.
Consequently, we recognize that developing a contactless sensing method to access these commands would significantly enhance drone localization and tracking performance.
However, the data is merely visible internally on the drone, and typical contactless, ground-based sensing methods cannot perceive it through external means.

\noindent \textbf{Propeller as Entry Point for Drone Sensing.}
In this work, we demonstrate that sensing internal control data using external sensors is feasible and transformative, achieved by monitoring propeller rotational speeds. 
The rationale behind this insight is that propeller speed crucially bridges internal commands and external status — it represents both the output of internal controls and influences the drone’s external flight status. 
By sensing this rotational speed, we could uncover flight control commands that were previously undetectable by existing drone-status sensing systems.

However, challenges still arise as drone propellers can rotate at 2,000 to 10,000 revolutions per minute (RPM) in during typical flights~\cite{deters2017static}. 
Traditional sensors, like cameras, radars, and acoustics, struggle with accurate rotational speed estimation due to limited spatio-temporal resolution~\cite{abdelnasser2023multirotor, kodukula2023squint, kodukula2022adaptive, garg2021owlet}. 
While sophisticated instruments like motion capture systems and laser velocimeters can measure such speeds, they necessitate deploying markers on propellers \cite{lin2022wi}, complicating robust, active, and contactless drone sensing systems.

Nowadays, event cameras, with their ultra-high-frequency and microsecond-level sampling, have revolutionized mobile applications such as high-speed motion tracking \cite{hao2023eventboost}, SLAM \cite{rebecq2016evo}, and drone obstacle avoidance \cite{xu2022swarmmap}. 
Unlike conventional cameras, event cameras asynchronously capture pixel-level intensity changes without global exposure times, eliminating motion blur even at extreme velocities \cite{wang2023ev, xu2023visible}. 
Their unique capabilities place them at the forefront of sensor technology, providing an innovative opportunity for real-time, high-speed propeller rotational speed estimation.

\noindent \textbf{Our work.}
We propose \textbf{\sysname}, a robust, high-precision, and low-latency system that explores drone dynamics through \textbf{Event}-camera-based \textbf{Pro}peller sensing. As shown in \fig\ref{fig:fig2}, with \sysname, the ultra-high-frequency rotational motion of propellers can be leveraged as a unique signature to $(i)$ infer drone \textit{internal flight commands} (\eg, hover/climb/descent, pitch/yaw/roll); and $(ii)$, based on $(i)$, estimate drone \textit{external flight status} with enhanced accuracy and efficiency. Notably, \sysname offers a complementary approach rather than a replacement for existing methods, introducing a novel sensing modality centered on propellers with validation in landing scenarios.

Making \sysname a practical system in the wild presents several challenges that are addressed in this work:
$(i)$ \textit{how to ensure the efficiency and accuracy of propeller sensing} given the immense data output of event cameras (\eg, millions of events per second) and their acute sensitivity to environmental variations, such as noise from drone vibrations and changes in lighting \cite{gallego2020event};
$(ii)$ \textit{how to infer both internal control commands and external flight status from the estimated propeller rotational speed.} 
This difficulty arises from the non-linear relationship between control commands and propeller speeds (\ie, reflecting generated force and torques) \cite{mahony2012multirotor}.
The design and implementation of \sysname excels in:

\noindent $\bullet$ \textbf{Count Every Rotation (\S\ref{sec:cer}).}
We propose an accurate and efficient event-camera-based propeller rotational speed estimation pipeline, including
$(i)$ a \textit{distribution-informed hierarchical event preprocessing} module to filter out background event noise and segment multiple propellers based on spatiotemporal statistics;
$(ii)$ a \textit{geometry-instructed event motion compensation} algorithm refines standard motion compensation \cite{gallego2018unifying} by incorporating propeller geometry information to accurately estimate rotational speed.
and $(iii)$ an \textit{adaptive event feature representation} strategy further enhances efficiency by adjusting and downsampling event batches based on motion consistency and spatiotemporal distribution.

\noindent $\bullet$ \textbf{Every Rotation Counts (\S\ref{sec:erc}).}
We demonstrate that leveraging propeller speed estimations will effectively profile drone dynamics from both internal and external perspectives. 
We first introduce a lightweight framework to predict drone actions by analyzing the rotational speed patterns of multiple propellers, that is, inferring \textit{internal control commands}. 
On this basis, we further treat the inferred control commands as a fresh, complementary modality to conventional sensors and propose a multi-modal fusion solution for enhancing drone \textit{external flight status} estimation performance.

We deployed \sysname with a Prophesee EVK4 HD event camera \cite{evk4} in various experimental testbeds and real-world drone delivery scenarios. 
Over 20 hours of experiments covered diverse robustness conditions, comparing \sysname's rotational speed estimation accuracy and latency against two baselines \cite{zhao2023ev,kolavr2024ee3p3d}. 
Results show that \sysname outperforms both baselines, reducing latency to 3.14$ms$ while achieving a merely 0.23\% estimation error, a >45\% error reduction.
\sysname also demonstrates its dual capabilities: it can accurately infer internal drone commands and enhance external flight status estimation. Specifically, it predicts drone flight actions with 96.5\% accuracy and then improves drone tracking precision by >22\% when integrated with existing sensing modalities.

In summary, this paper makes the following contributions:

\noindent (1) We rethink active, contactless, and ground-based drone sensing by expanding our focus from specific sensors or algorithms to include the entire drone flight control system. 
We demonstrate that centering on propeller rotational speed enables the system to capture both internal flight control commands and external flight status, introducing a new dimension to drone sensing problems.

\noindent (2) We introduce \sysname, a comprehensive solution utilizing event cameras for high-precision, real-time measurement of multi-propeller speeds, \ie, \textit{count every rotation.}
Based on the propeller speed, we further propose algorithms to accurately estimate both the internal and external flight status of drones, that is, \textit{every rotation counts.}

\noindent (3) We fully deploy \sysname in both experimental testbeds and real-world drone delivery scenarios.
Extensive evaluations demonstrate its superior performance. 
\textit{Artifacts will be made publicly available before publication.}

\section{System Overview}
\begin{figure}
    \setlength{\abovecaptionskip}{0.2cm} % height above Figure X caption
    \setlength{\belowcaptionskip}{-0.3cm}
    \centering
    \includegraphics[width=0.98\columnwidth]{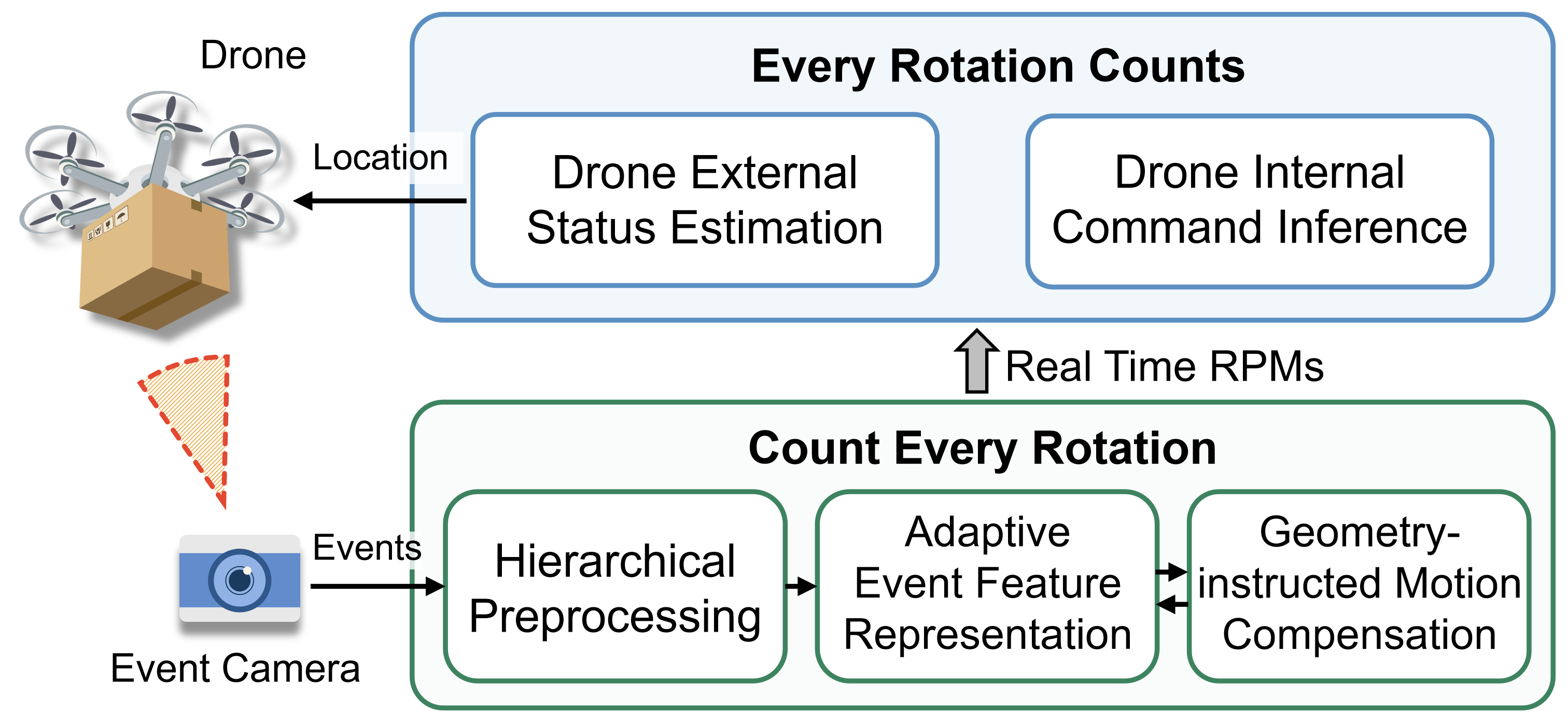}
    \caption{System architecture of \sysname.}
    \label{fig:overview}
    \vspace{-3.6mm}
\end{figure}

From the top perspective, we design and implement \sysname, a cutting-edge, high-precision, and low-latency system that explores drone dynamics through event-camera-based propeller sensing. 
\fig \ref{fig:overview} illustrates the architecture of \sysname, which provides system support on two aspects.
% \vspace{-0.5cm}

\noindent\textbf{Count Every Rotation}. 
\sysname consists of three key components for measuring the propeller rotation speed.
Specifically, \sysname begins by employing the \textit{Distribution-informed Hierarchical Event Preprocessing} ($\S$\ref{preprocess}) module to filter out noise and segment event streams for each propeller. Based on the segmented results, the \textit{Adaptive Event Feature Representation} ($\S$\ref{ASER}) module progressively establishes an adaptive event batch based on the rotational motion consistency of events. Finally, leveraging this adaptive event batch, the \textit{Geometry-instructed Event Motion Compensation} ($\S$\ref{GEMC}) module employs a motion compensation framework with a tailored objective function to optimize motion parameters, providing accurate rotational speeds for all propellers.

\noindent\textbf{Every Rotation Counts}. Based on the real-time propeller speed estimations, \sysname effectively profiles drone dynamics from both internal and external perspectives:\\
\noindent $\bullet$ \textit{Drone Internal Command Inference} ($\S$\ref{sec:intention}). We design a lightweight framework to infer drone internal commands by analyzing the distinct patterns encoded in the rotational speed distribution of the drone's multiple propellers.\\
\noindent $\bullet$ \textit{Drone External Status Estimation} ($\S$\ref{sec:localization}). We design a multi-modal fusion algorithm to enhance drone external status estimation performance by leveraging the propeller speed information as a complementary motion prior and integrating it with existing sensing modalities.

\section{Count Every Rotation}\label{sec:cer}
\begin{figure}[t]
    \setlength{\abovecaptionskip}{0.1cm} % height above Figure X caption
    \centering
    \includegraphics[width=0.99\columnwidth]{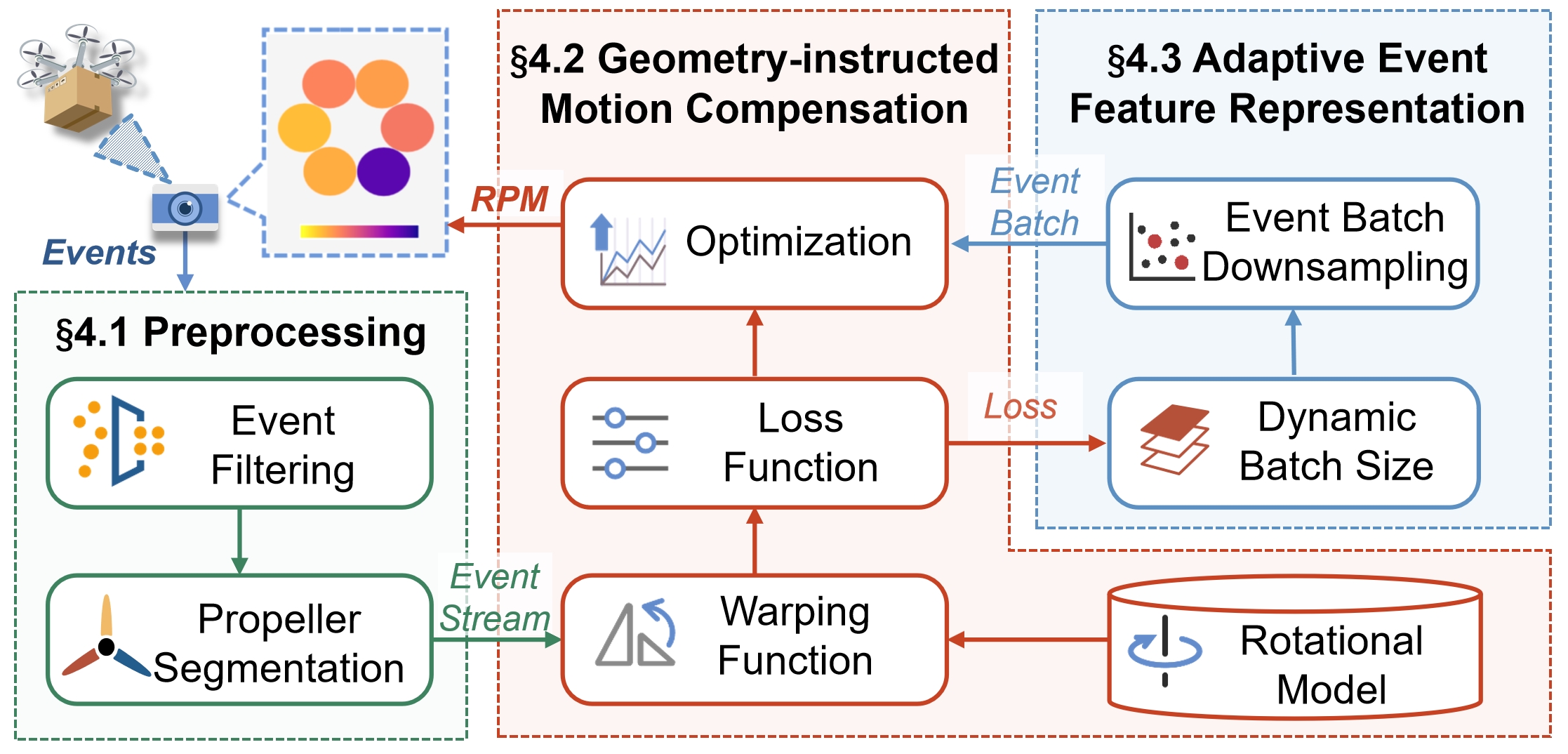}
    \caption{Workflow of \textit{Count Every Rotation}.}
    \label{fig:cer}
    \vspace{-3.0mm}
\end{figure}

In this section, we begin with an overview of event cameras and their challenges. We then detail the design of our event preprocessing method (\S \ref{preprocess}) and the propeller speed estimation algorithms for high accuracy (\S \ref{GEMC}) and efficiency (\S \ref{ASER}).
The algorithm workflow is shown in \fig \ref{fig:cer}.

\begin{figure}[t]
    \setlength{\abovecaptionskip}{-0.1cm} % height above Figure X caption
    \setlength{\belowcaptionskip}{-0.4cm}
    \centering
        \subfigure[Event generation principle]{
            \centering
            \includegraphics[width=0.47\columnwidth]{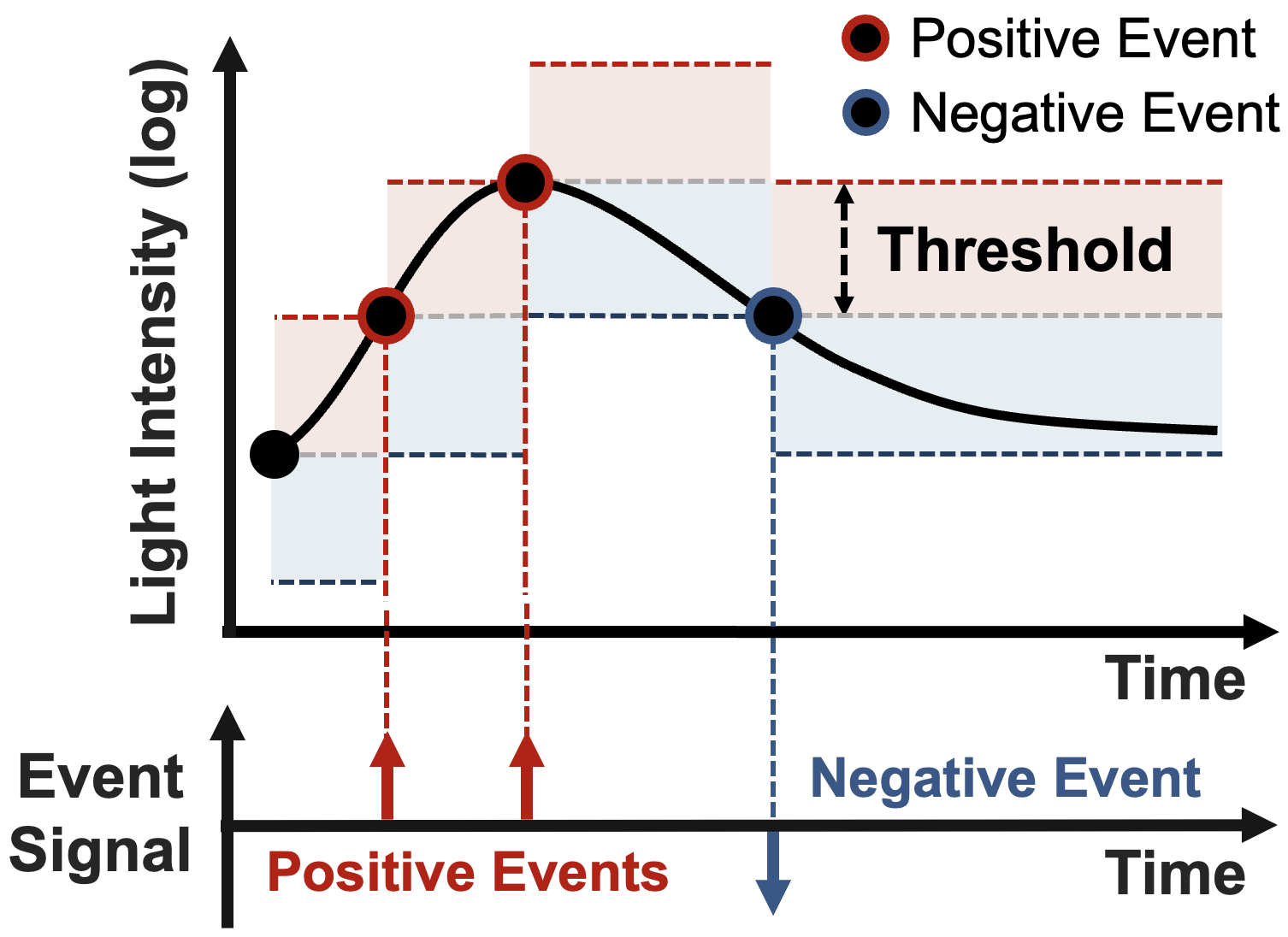}
            %\caption{fig1}
        }%
        \hspace{0cm} 
        \subfigure[Comparing event generating speed]{
            \centering
            \includegraphics[width=0.47\columnwidth]{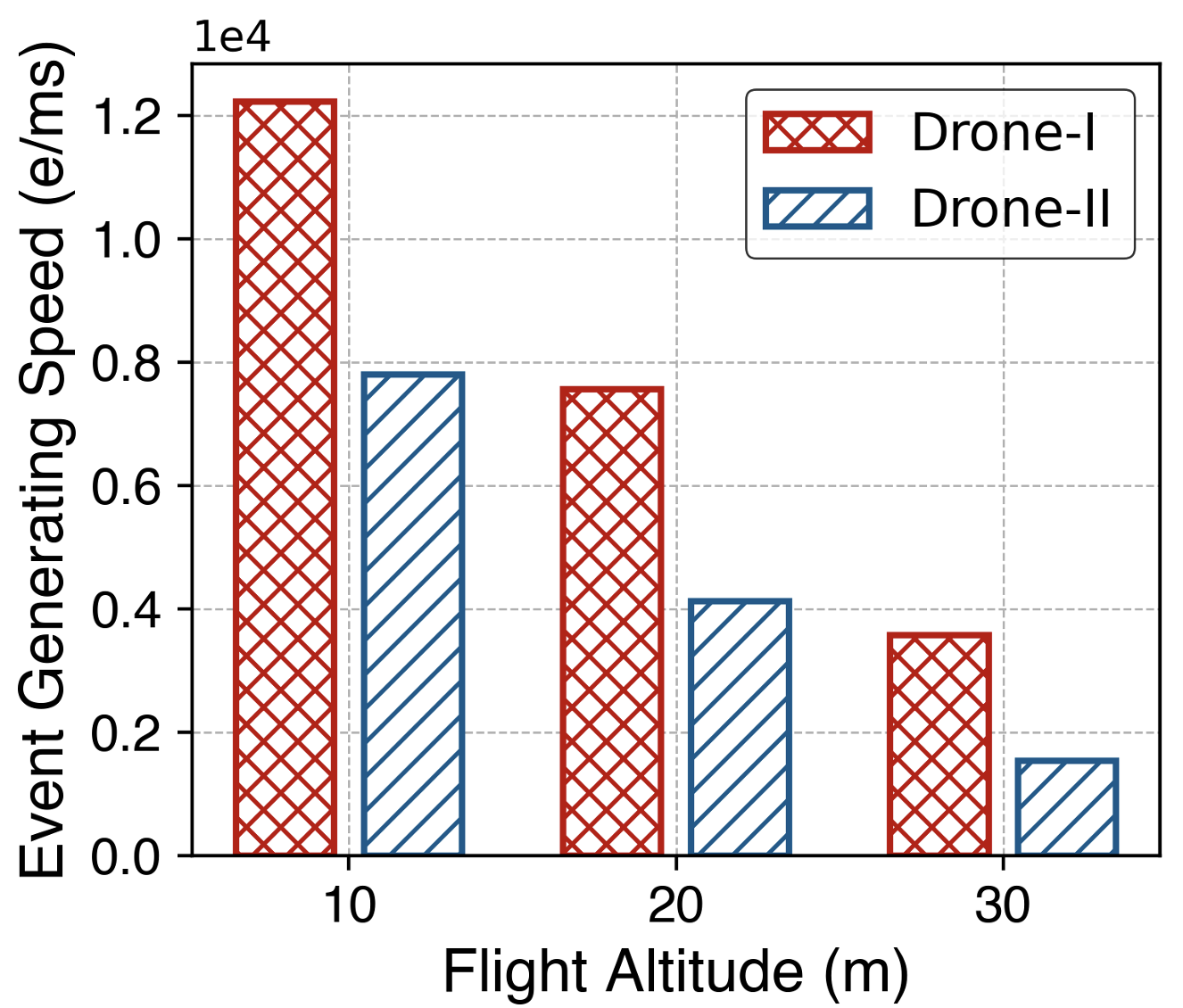}
            %\caption{fig1}
        }%
    \caption{(a) Event generation principle. (b) The high event-generating speed of two typical commercial drones (Drone-I\cite{m30t} and Drone-II\cite{meituan}) even hovering captured by a commercial event camera.}
    \label{fig:bg}
    \vspace{-2mm}
\end{figure}

Event cameras are bio-inspired sensors that operate independently and asynchronously for each pixel. Instead of capturing images at a fixed time interval, event cameras record asynchronous changes in pixel brightness, resulting in a stream of events with \textit{microsecond} resolution \cite{gallego2020event}, as shown in \fig \ref{fig:bg}a. 
When a pixel detects a light intensity change exceeding a predefined magnitude, it immediately outputs an event $e_k \doteq \left(\mathbf{x}_k, t_k, p_k\right)$, encoding the triggered position $\mathbf{x}_k$, the triggered time $t_k$, and polarity $p_k$ (+1 for brighter and -1 for darker) of the intensity changes.

The high temporal resolution empowers event cameras to estimate real-time propeller rotational speed. However, translating \sysname into a practical system in the wild faces two fundamental challenges.

\noindent $\bullet$ \textit{High sensitivity to scene dynamics impairs speed estimation.} 
During operation, the generated noisy events include random background noise (\eg, hot noise, $1/f$ noise\cite{gallego2020event}) and those triggered by vibrations of the drone's body and propellers. These noisy events often intermix within the space-time events triggered by propeller rotational motion (\fig \ref{fig:preprocess}b), leading to significant interference in estimation. 

\noindent $\bullet$ \textit{High-frequency sensing data induces intensive computational resource overhead.} 
As shown in \fig \ref{fig:bg}b, an event camera observing a typical hovering drone rotating at approximately 3000 RPM can generate over $10^4$ events per millisecond. 
Processing this high volume of events is computationally intensive and frequently leads to substantial delays.

\subsection{Distribution-informed Hierarchical Event Preprocessing} \label{preprocess}
Event cameras generate a dense stream of events in short periods, which includes data from multiple rotating targets as well as background noise. Consequently, it is essential to perform event filtering and segment events from different propellers before conducting further analysis.

\subsubsection{\textbf{Temporal Distribution-informed Event Filtering}}
\begin{figure*}[t]
    \setlength{\abovecaptionskip}{0.1cm} % height above Figure X caption
    \setlength{\belowcaptionskip}{-0.2cm}
    \centering
    \includegraphics[width=1.96\columnwidth]{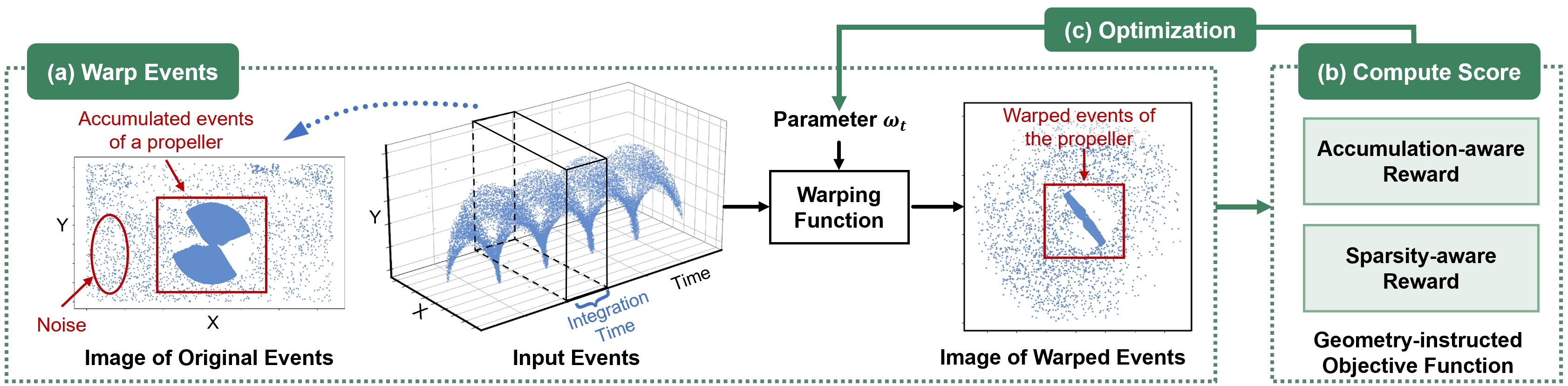}    
    \caption{Illustration of the geometry-instructed event motion compensation scheme.}
    \label{fig:cmax}
    % \vspace{-2mm}
\end{figure*}
\begin{figure}[t]
    \setlength{\abovecaptionskip}{0.0cm} % height above Figure X caption
    \setlength{\belowcaptionskip}{-0.2cm}
    \centering
    
    % 第一行
    \subfigure[Gray Image]{
        \centering
        \includegraphics[width=0.46\columnwidth]{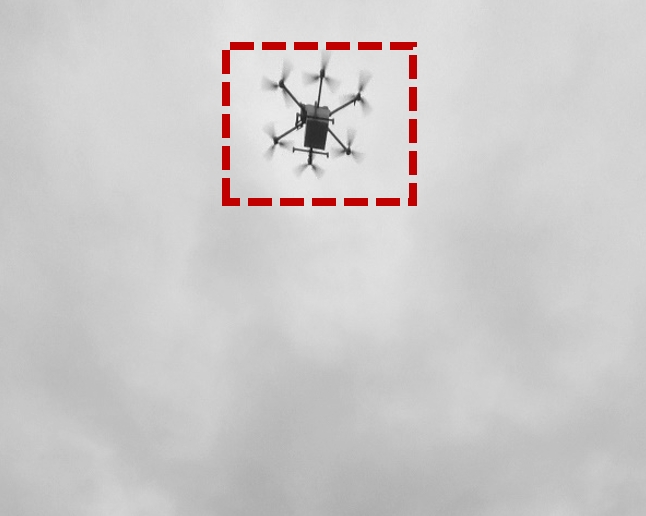}
    }%
    \hspace{0cm} 
    \subfigure[Original Event Stream]{
        \centering
        \includegraphics[width=0.46\columnwidth]{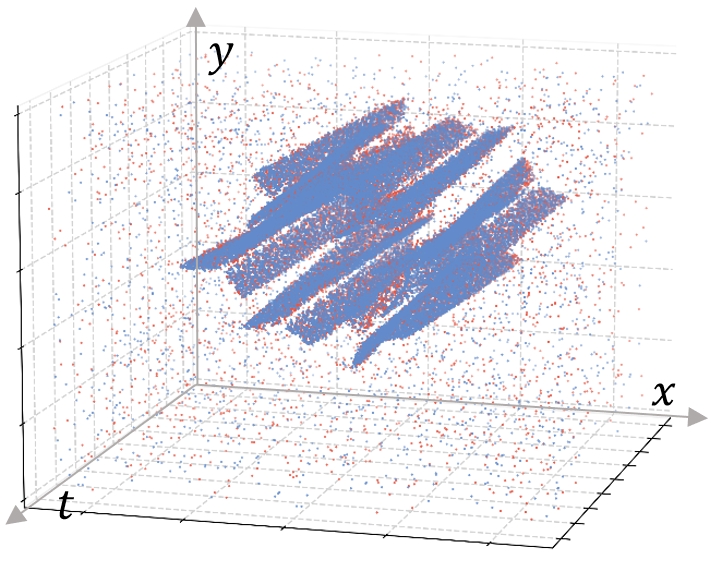}
    }%
    % 换行∏
    \vspace{0.1cm}
    % 第二行
    \subfigure[Filtered Event Stream]{
        \centering
        \includegraphics[width=0.46\columnwidth]{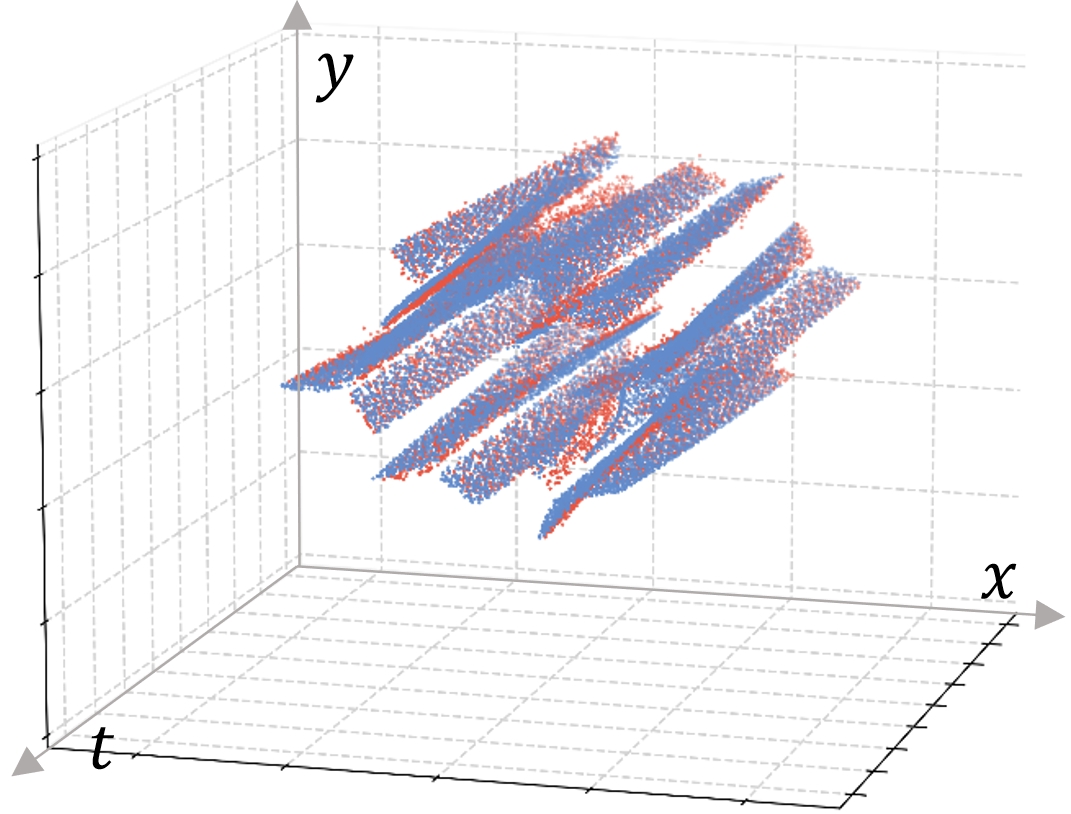}
    }%
    \hspace{0cm} 
    \subfigure[Segmentation Performance]{
        \centering
        \includegraphics[width=0.46\columnwidth]{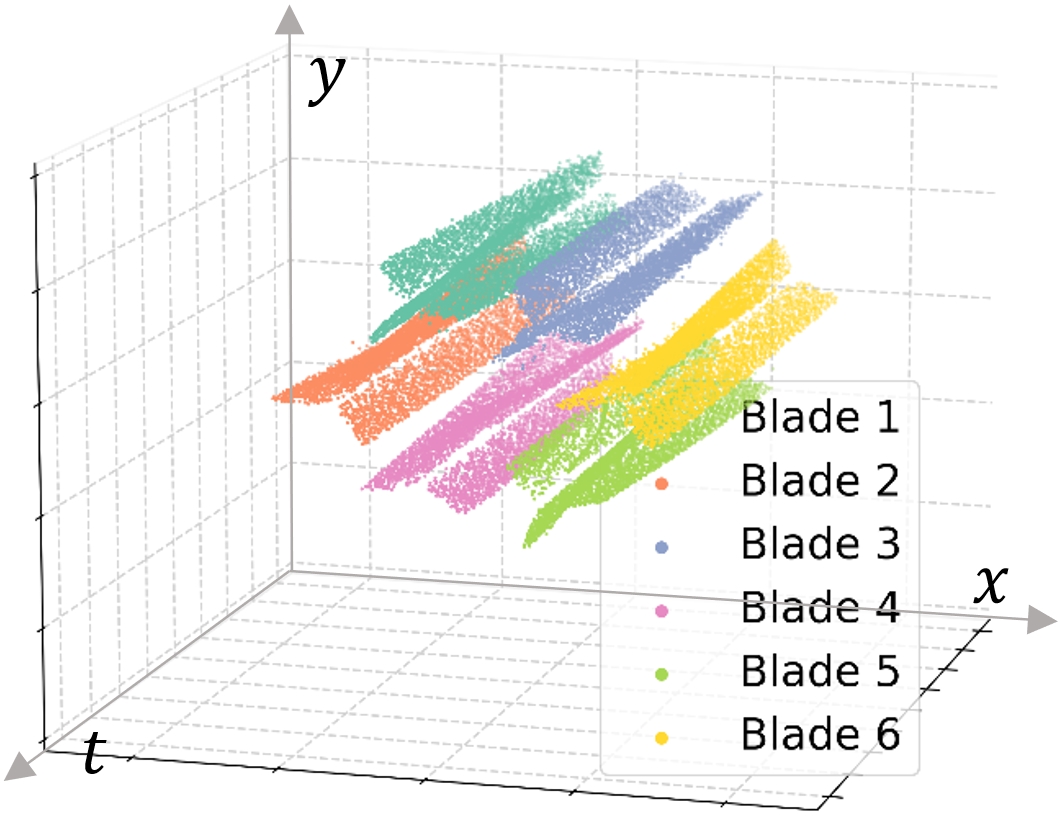}
    }%
    
    \caption{Step-by-step event preprocessing illustration.}
    \label{fig:preprocess}
    \vspace{-4mm}
\end{figure}
We transforms the temporal distribution of events into heatmaps to filter noise using statistical metrics. Specifically, for events collected within the time window $[t_o, t_0+\delta t]$, they are binned into a 2D histogram based on their coordinates, with each bin representing a specific spatial region (e.g., $5 \times 5$ pixels). 
The characteristics of each bin are usually inconsistent in two aspects:
($i$) \textit{Event counts:} high-speed rotational motion generates more events than noise. Bins with event counts significantly below the mean (e.g., three times lower) likely contain noise.
($ii$) \textit{Event polarities:} rotational motion produces bipolar events, whereas background noise tends to be unipolar \cite{xu2024seeing}. Bins with highly unbalanced distributions of positive and negative events indicate noise.

Inspired by these metrics, we create two heatmaps that represent event counts in each bin and the proportion of positive events, respectively. Rotation-triggered events exhibit higher event counts and a more balanced proportion of event polarities (closer to 1:1). \sysname utilizes these patterns to effectively filter out noise.

\subsubsection{\textbf{Spatial Distribution-informed Propeller Segmentation}}
Next, we segment events from different propellers using the K-means clustering algorithm, which identifies $K$ centroids representing the data, with each data point assigned to the nearest centroid to form clusters. 
Assuming a drone with $K$ propellers and an event stream $\mathcal{E}_s=\{e_l\}^{N_s}_{l=1}$, our goal is to isolate events associated with each rotating propeller. Therefore, the objective function is $\min _{\mathcal{C}_k, \mathbf{m}_k} \sum_{k=1}^K \sum_{\mathbf{x}_l \in \mathcal{C}_k}\left\|\mathbf{x}_l-\mathbf{m}_k\right\|_2^2$,
where $\mathbf{x}_l$ is the spatial coordinate of event $e_l$,  $\mathcal{C}_1, \ldots, \mathcal{C}_K$ are $K$ disjoint clusters, and $\mathbf{m}_k=\sum_{\mathbf{x}_l \in \mathcal{C}_k} \mathbf{x}_l /\left|\mathcal{C}_k\right|$ is the centroid of $\mathcal{C}_k$. This function minimizes the within-cluster sum of squared errors, effectively maximizing cluster separation.

By optimizing this objective, we assign each event to a specific propeller, with the determined centroids representing the rotational centers of the propellers. This method enables accurate and efficient segmentation of events for subsequent rotational speed estimation and analysis. The step-by-step event preprocessing results are depicted in \fig \ref{fig:preprocess}.

\subsection{Geometry-instructed Event Motion Compensation}\label{GEMC}

Although preprocessing filters out some noise with distinct features, stubborn residues persist since: ($i$) the sources of noise are multiple (\eg, background activity noise, hot noise, and $1/f$ noise) and their firing time is random~\cite{gallego2020event}; and ($ii$) vibrations from the drone's body and propellers triggers additional events. Consequently, this noisy data is interwoven in the space-time volume, making it difficult to fully filter out and hindering reliable motion estimation.

To address this issue, we introduce a geometry-instructed motion compensation pipeline. The basic idea is that, as a propeller edge moves on the image plane, it triggers events on the pixels it traverses. Thus the motion of the propeller can be estimated by warping the events to a reference time and maximizing their alignment. We design an objective function to optimize this alignment performance based on two key insights from propeller geometry:
(\textit{i}) If the correct parameter is found and events are warped accordingly, most events will accumulate at the propeller locations at the reference timestamp;
(\textit{ii}) The best parameter will produce the minimum number of locations containing events, highlighting extreme ``sparsity'' on the image plane.

This design not only implicitly manages data association without costly computation, but also enables accurate estimation free from noise interference. We first introduce the motion compensation framework, and then detail its warping function and the objective function components.

\subsubsection{\textbf{Motion Compensation Framework}}\label{sec:cmax}
Given an event batch $\mathcal{B}=\{e_k\}^{N_e}_{k=1}$ of a rotating propeller, our motion compensation algorithm comprises three stages (\fig \ref{fig:cmax}):

\textit{(a)} Warp the events to a reference timestamp. 
We warp the events backward to the reference timestamp $t_{ref}$, \ie, $e_k \doteq \left(\mathbf{x}_k, t_k, p_k\right) \mapsto\left(\mathbf{x}_k^{\prime}, t_{\mathrm{ref}}, p_k\right) \doteq e_k^{\prime}$ according to the warping function $\mathbf{x}_k^{\prime} = \mathbf{W}(\mathbf{x}_k, t_k;\omega)$, where $\omega$ is the candidate motion parameter. 
% where the warping function $\mathbf{W}$ indicates the coordinate transformation.
Intuitively, this stage can be seen as an inverse process of the rotation process of the propeller. 

\textit{(b)} Compute a score $R$ based on the image of warped events. 
% In this key step of the motion compensation framework, 
We first build an image $H$ of warped events, $H(\mathbf{x} ; \omega) \doteq \sum_{k=1}^{N_e} \delta\left(\mathbf{x}-\mathbf{x}_k^{\prime}\right)$,
% \begin{equation}
% \vspace{-0.2cm}
% H(\mathbf{x} ; \omega) \doteq \sum_{k=1}^{N_e} \delta\left(\mathbf{x}-\mathbf{x}_k^{\prime}\right),
% \end{equation}
where each pixel $\mathbf{x}$ sums the count of warped events $\mathbf{x}_k^{\prime}$ that fall within it (indicated by the Dirac delta $\delta$). 
Next, we create an objective function $R(H(\mathbf{x}; \omega))$ to evaluate the fitness of the candidate parameter $\omega$, detailed in \sec \ref{sec:obj}.

\textit{(c)} Optimize the objective function with respect to the parameter $\omega$. Once the objective function is constructed, various optimization tools can be utilized. Note that our algorithm is flexible and not dependent on a specific optimizer. In this work, we apply Brent's method \cite{press1992numerical} to obtain the best model parameters.

\subsubsection{\textbf{Warping Function}}
\begin{figure}[t]
    \setlength{\abovecaptionskip}{0.0cm} % height above Figure X caption
    \setlength{\belowcaptionskip}{-0.2cm}
    \centering
        \subfigure[Propeller rotational mode]{
            \centering
            \includegraphics[width=0.47\columnwidth]{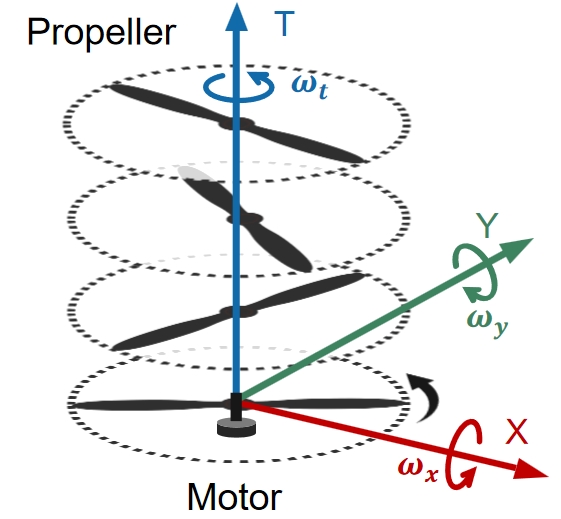}
            %\caption{fig1}
            % \label{fig:indoor_acc}
        }%
        \hspace{0cm} 
        \subfigure[Generated event stream]{
            \centering
            \includegraphics[width=0.47\columnwidth]{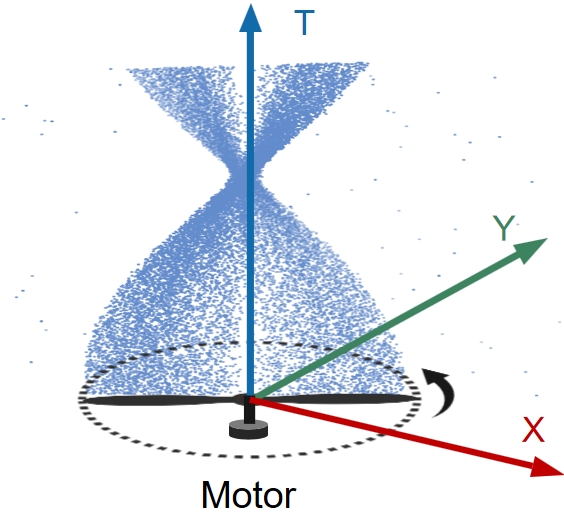}
            %\caption{fig1}
            % \label{fig:indoor_acc}
        }%
    \caption{Illustration of the propeller rotational model.}
    \label{fig:model}
    % \vspace{-4mm}
\end{figure}

Recall that each event data includes spatial and temporal information.
We first build a coordinate $OXYT$ fixed relative to the propeller, with the origin at the rotor center (\fig \ref{fig:model}a). The event set can then be viewed as a set of 3D points in this space (\fig \ref{fig:model}b). 
The motion parameters $\omega=(\omega_x, \omega_y, \omega_t)$ represent the angular velocities around the $(x,y,t)$ axes.

Consider a rotation matrix $\mathrm{R}(t_k;\omega)$ that describes the 3D rotation motion at $t_k$. This matrix can be decomposed into three independent rotations—roll, pitch, and yaw—around each axis, $\mathrm{R}=\mathrm{R}_X(\omega_x) \mathrm{R}_Y(\omega_y) \mathrm{R}_T(\omega_t)$,
where $\mathrm{R}_X, \mathrm{R}_Y, \mathrm{R}_T$ are the rotation matrices of three axes, respectively \cite{euler}. 
For a short period, we assume the rotational angles in the $X$ and $Y$ dimensions remain constant and events are generated mostly by rotational motion around $T$ axis (as shown in \fig \ref{fig:model}b). 
For instantaneous rotational speed estimation over a short time period, we assume the rotational angles about the $X$ and $Y$ axes remain constant and events are generated mostly by rotational motion around $T$ axis (as shown in \fig \ref{fig:model}b). Thus the corresponding rotation matrix $\mathrm{R}$ is given by: 
\begin{equation}
    \mathrm{R}(\omega_t)=\left[\begin{array}{ccc}
\cos (\omega_t) & \sin (\omega_t) & 0 \\
-\sin (\omega_t) & \cos (\omega_t) & 0 \\
0 & 0 & 1.
\end{array}\right]
\label{eq:model}
\end{equation}

Given a reference timestamp $t_{ref}$, as the propeller rotates, the event points transform according to the rotational model:
\begin{equation}
\begin{aligned}
    \overline{\mathbf{x}}(t_k) & \propto \mathrm{R}(t_k-t_{ref};\omega_t)\overline{\mathbf{x}}(t_{ref})\\
    &=\exp(\hat{\omega_t}(t_k-t_{ref}))\overline{\mathbf{x}}(t_{ref}),
\end{aligned}
\end{equation}
where $\overline{\mathbf{x}} \propto (\mathbf{x}^T,1)^T$ are homogeneous coordinates;  $\mathrm{R}(t_k-t_{ref};\omega_t)$ is the rotation matrix of the rotational motion, $\exp$ is the exponential map of the rotation group $SO(3)$ and $\hat{\omega_t}$ is the cross-product matrix associated to $\omega_t$. 
Therefore, the warping function can be derived as:
\begin{equation}
\begin{aligned}
        \mathbf{x}_k^{\prime} 
        &= \mathbf{W}(\mathbf{x}_k, t_k;\omega_t)
        \propto \mathrm{R}^{-1}(t_k-t_{ref};\omega_t)\overline{\mathbf{x}}_k,
        % &\propto \exp(-\hat{\omega_t}(t_k-t_{ref}))\overline{\mathbf{x}}_k,
\end{aligned}
\end{equation}

\begin{figure}[t]
    \setlength{\abovecaptionskip}{0.0cm} % height above Figure X caption
    \centering
        \subfigure[Poorly-parameterized warping]{
            \centering
            \includegraphics[width=0.47\columnwidth]{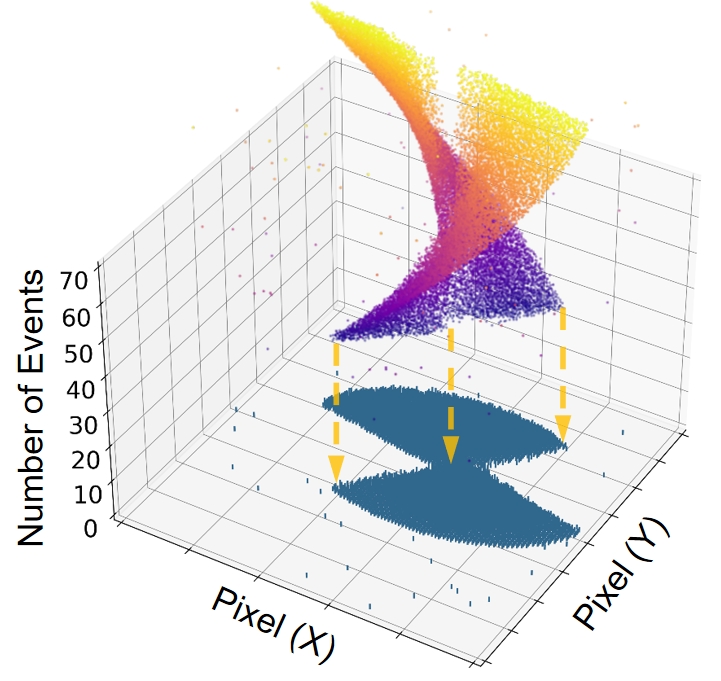}
            %\caption{fig1}
            % \label{fig:indoor_acc}
        }%
        \hspace{0cm} 
        \subfigure[Well-parameterized warping]{
            \centering
            \includegraphics[width=0.47\columnwidth]{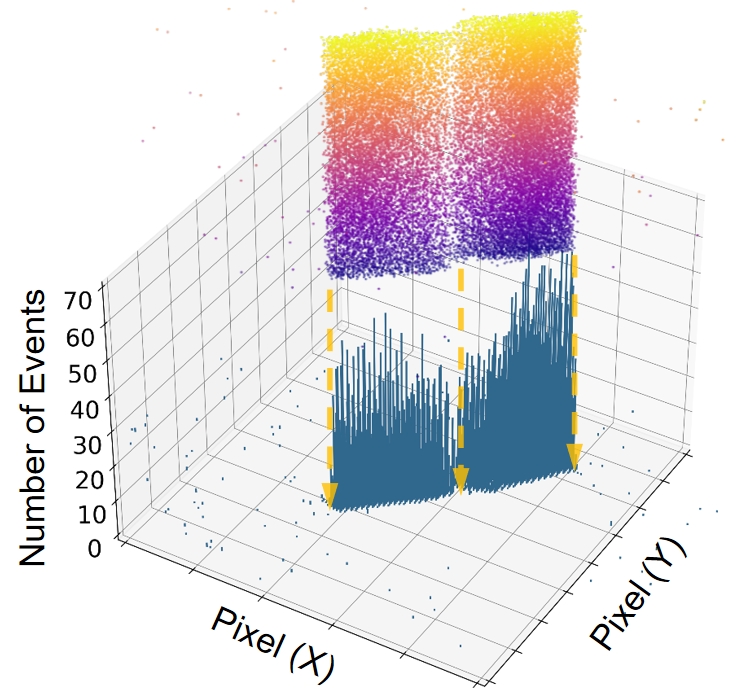}
            %\caption{fig1}
            % \label{fig:indoor_acc}
        }%
    \caption{Comparison of poor (a) and accurate (b) parameter estimates. \textnormal{Two figures show varying event accumulations on the image plane with different estimation accuracy. Event colors represent different times.}}
    \label{fig:magloss}
    \vspace{-4mm}
\end{figure}
\begin{figure}[]
    \setlength{\abovecaptionskip}{-0.0cm} % height above Figure X caption
    \setlength{\belowcaptionskip}{-0.2cm}
    \centering
    \includegraphics[width=1.0\columnwidth]{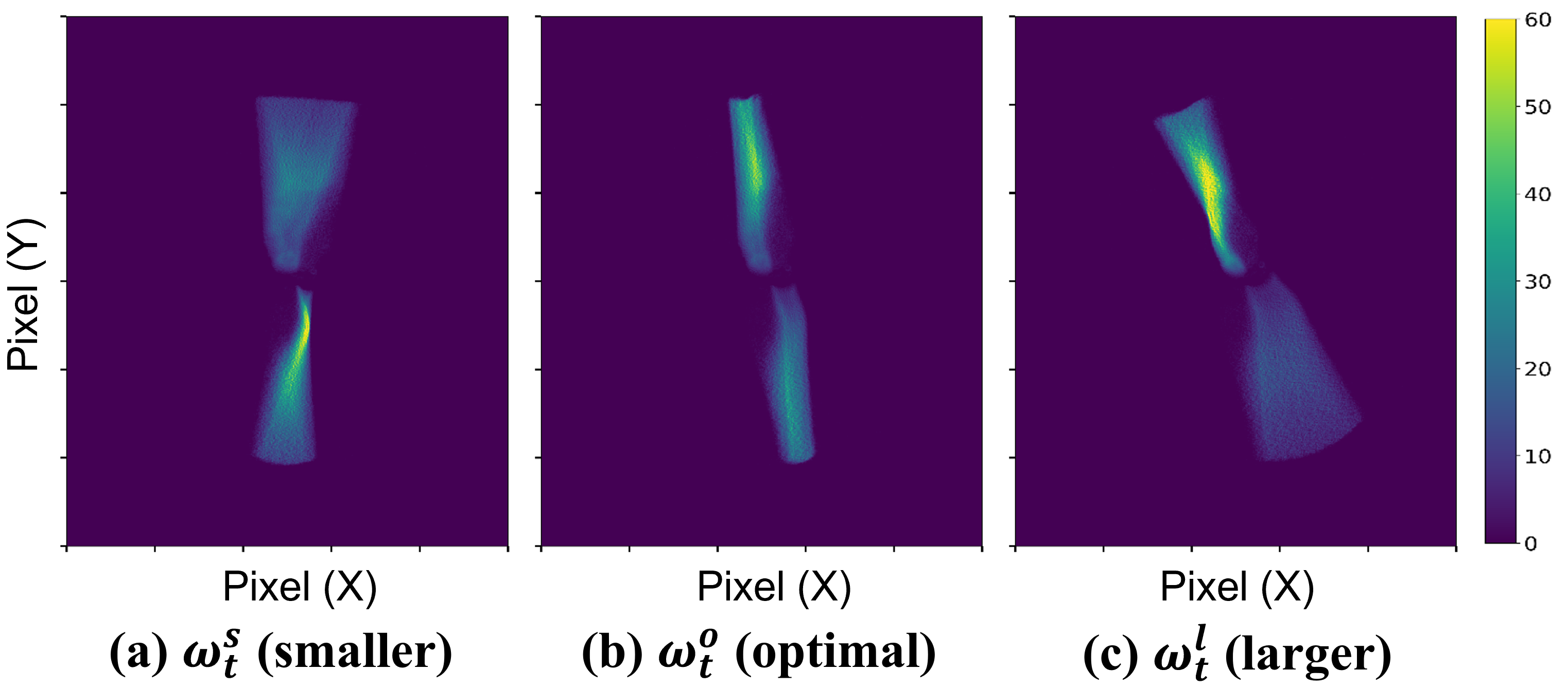}
    % \caption{The magnitude-aware reward for different candidate parameters $\omega_t$: (a) a bit smaller than ground truth, (b) ground truth, (c) a bit larger than ground truth.}
    \caption{Illustration of heatmaps of warped events for various parameters $\omega_t$. 
    \textnormal{At the optimal parameter, the image patch exhibits the highest ``sparsity'', with the minimum number of locations showing events.}}
    \label{fig:sparloss}
    \vspace{-2mm}
\end{figure}
\subsubsection{\textbf{Geometry-instructed Objective Function}}\label{sec:obj}
As mentioned above, we aim to build an objective function to evaluate the quality of the warped image with respect to the parameter $\omega_t$. This objective function will guide the following optimization process. Therefore, the best aligned warped image is desired where the events triggered by the rotational motion are correctly warped to the locations at the reference timestamp, while the estimated trajectory is not influenced by the noise event data. To evaluate the quality of the warped image with respect to the parameter $\omega_t$, we design a dual-reward objective function by incorporating the propeller geometry, as elaborated below.

\noindent \textbf{Accumulation-aware reward}.
Given that events triggered by rotational motion follow similar point trajectories, whereas noisy events do not, warping these events according to the best parameter should ideally result in the accumulation of most events at the propeller's initial locations.
As shown, the well-warped events in \fig \ref{fig:magloss}b result in a high accumulation in the XY plane, while the poorly-warped events in \fig \ref{fig:magloss}a with the same total accumulations spread across many locations.

Motivated by this observation, \sysname rewards the large accumulation of events at some locations on the image plane, 
which can be expressed as:
\begin{equation}
    r_{\mathrm{acc}}(H(\mathbf{W}(\mathbf{x}_k, t_k;\omega_t)))=\sum_{i, j} \exp (h(i, j)),
\end{equation}
where $h(i,j)$ is the value of pixel $(i,j)$ in $H$. 
This design gives locations with many accumulations in them a higher value, breaking away from noise interference.

\begin{figure*}[t]
    \setlength{\abovecaptionskip}{0.0cm} % height above Figure X caption
    \setlength{\belowcaptionskip}{-0.2cm}
    \centering
        \subfigure[Illustration of consistency-driven dynamic batch size.]{
            \centering
            \includegraphics[width=1.17\columnwidth]{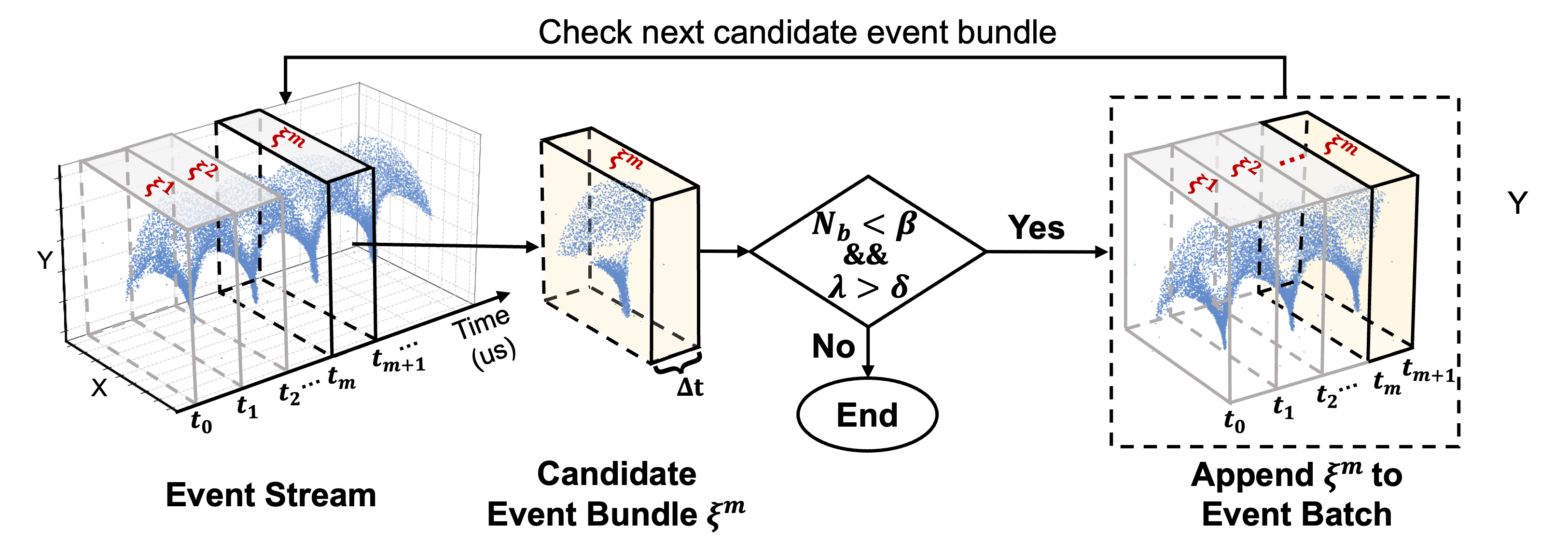}
            %\caption{fig1}
            % \label{fig:indoor_acc}
        }%
        \hspace{2mm} 
        \subfigure[Illustration of structure-aware downsampling]{
            \centering
            \includegraphics[width=0.77\columnwidth]{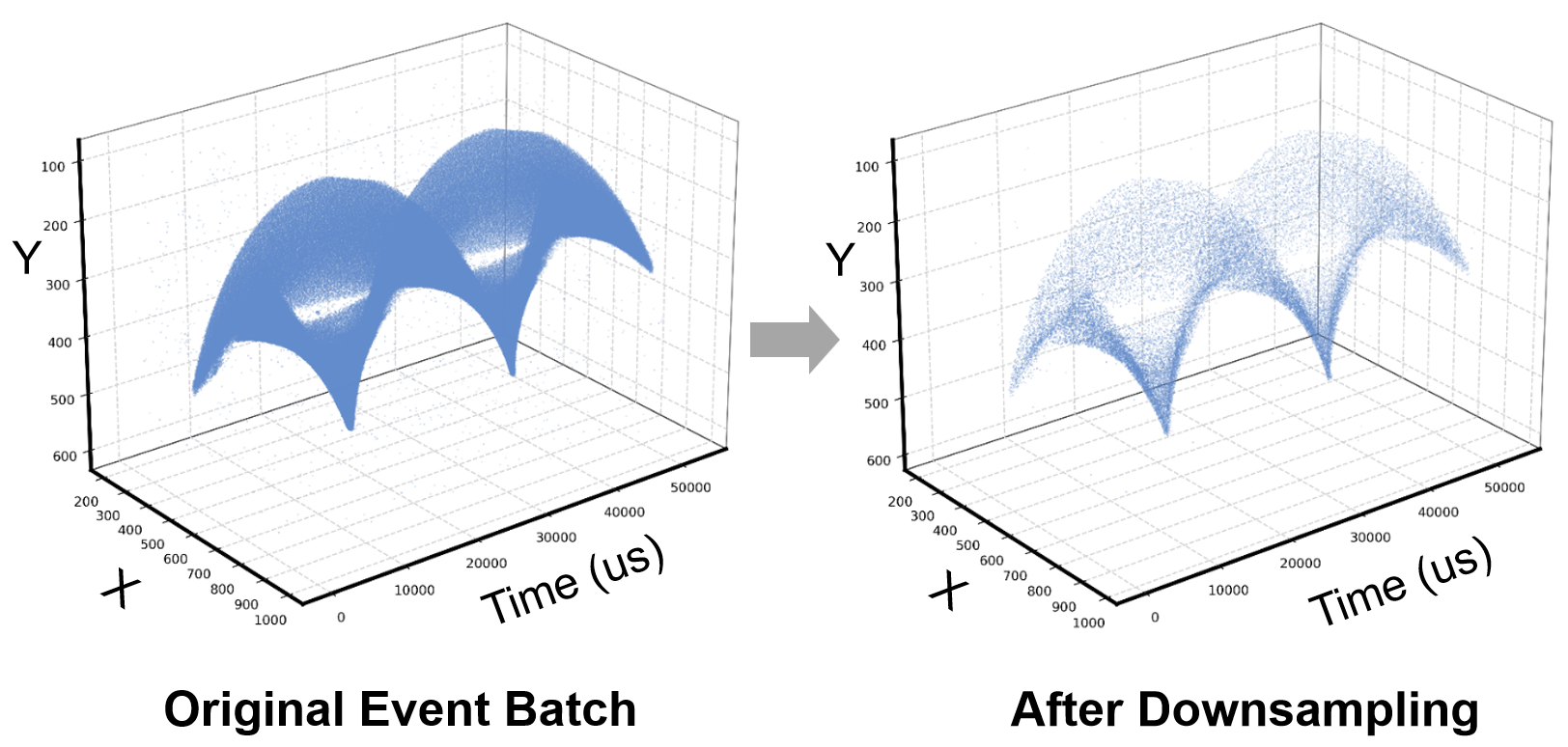}
            %\caption{fig1}
            % \label{fig:indoor_acc}
        }%
    \caption{Illustration of adaptive event feature representation.}
    \label{fig:batch}
    % \vspace{-4mm}
\end{figure*}
\noindent \textbf{Sparsity-aware reward}.
\fig \ref{fig:sparloss} presents the heatmaps of warped events with three candidate parameters $\omega_t$. As seen, parameters slightly larger or smaller than optimal can achieve similar or even greater event accumulations (brighter color), promoting these nearly correct but less accurate estimations. On the other hand, the image of warped events with the optimal parameter shows the most ``sparsity'', which is consistent with the better alignment of the warped events.

Inspired by this observation, we propose a simple yet effective sparsity-rewarding function to promote image sparsity:
\begin{equation}
    r_{\mathrm{spa}}(H(\mathbf{W}(\mathbf{x}_k, t_k;\omega_t)))=\sum_{i, j} \frac{1}{\exp(h(i, j))-1+\epsilon}.
\end{equation}
This function essentially encourages fewer locations containing events, making it more sensitive to different trajectories and thereby enabling more accurate estimation.

Finally, we build the objective function by linearly combining the accumulation-aware and sparsity-aware rewards, which can be expressed as, $R(H(\mathbf{W}(\mathbf{x}_k, t_k;\omega_t)))=r_{\mathrm{acc}}(H)+r_{\mathrm{spa}}(H)$. 
% This objective function integrates the complementary advantages of both kinds of reward, promoting more precise, noise-tolerant, and lightweight estimation.
This objective function integrates the complementary advantages of both rewards, promoting precise and noise-tolerant estimation.

\subsection{Adaptive Event Feature Representation}\label{ASER}
Event cameras generate large data volumes quickly, causing processing delays with increased accumulation time. Variations in device specifications (e.g., resolutions) and target states (e.g., scene dynamics, textures, spatial locations) cause the data volume to vary greatly, from hundreds to tens of thousands of events per millisecond. Current practices often overlook these variations \cite{xu2024seeing, falanga2020dynamic}, utilizing fixed batch sizes or fixed time intervals for event accumulation, which leads to information loss or high computational burdens.

To address these issues, we explore the fundamental principles of event cameras and ground our design on two key insights:
$(i)$ variations in rotational speed impact the motion consistency encoded in the events;
and $(ii)$ events triggered during a very short period follow the same motion pattern, reducing the necessity to consider all events for estimation.

Based on these basic ideas, we propose an adaptive event feature representation that includes:
(\textit{i}) a consistency-driven dynamic batch size strategy to adaptively accumulate events, ensuring the constant velocity assumption holds within each batch without compromising estimation accuracy;
and (\textit{ii}) an effective downsampling algorithm to select a representative subset of event points, accelerating event data association while preserving informative spatial features.

\subsubsection{\textbf{Consistency-driven Dynamic Batch Size}}
We propose a consistency-driven dynamic batch size strategy to control the amount of events to be accumulated for each estimation step, which is more generalizable to different scene textures, rotational speeds, and camera resolutions compared to the fixed batch strategy. 
First, we divide the event stream $\mathcal{E}$ of duration $T$ into multiple event bundles by a constant time interval $\Delta t$, and each event bundle $\xi^m = \left\{e_i \mid \tau_m \leq t_i \leq \tau_{m+1}\right\}$ contains events within the time period $\left[\tau_m, \tau_{m+1}\right]$, where $\tau_{m+1}=\tau_m+\Delta t$, $0 \leq m < T/\Delta t$.
Next, we create an event batch $\mathcal{B}$ with the first event bundle $\xi^0$ and initialize the motion parameter $\omega_t$. 
For each subsequent bundle, we check the following two criteria.

\noindent \textbf{Bundle consistency criterion}.
We measure the consistency of bundles by a factor, namely consistency rate $\lambda \in [0,1]$. It is computed by warping the candidate event bundle $\xi^m$ and the last bundle $\xi^k$ in the current event batch to the same reference timestamp with the current parameter, and comparing their objective function values:
\begin{equation}
\begin{aligned}
\lambda &= \left| \frac{R(H(\mathbf{x}_m,t_m;\omega_t)) - R(H(\mathbf{x}_k,t_k;\omega_t))}{R(H(\mathbf{x}_k,t_k;\omega_t))} \right|.
\end{aligned}
\end{equation}

Intuitively, a high $\lambda$ value indicates a high probability that the current bundle has a different rotational speed compared with the existing events in the batch. Therefore, we set a threshold $\delta$ (\eg, 0.3) for the determination. If the consistency rate $\lambda$ of the candidate bundle $\xi^m$ is lower than $\delta$, we append this new bundle to the current event batch. Otherwise, we stop appending bundles and update the motion parameter $\omega_t$ based on the final event batch.

\noindent \textbf{Bundle number criterion}.
Assuming an ideal scenario in which the consistency rate remains below the threshold, the number of events in the current batch will increase as more bundles are added. While this growth enhances the reliability of speed estimation, the marginal benefit will diminishes and in turn increases computational demand. Therefore, the appending process will also terminate if the total bundle number $N_b$ contained in the event batch reaches a certain threshold $\beta$. The overall workflow is illustrated in \fig \ref{fig:batch}a.

\begin{figure*}[t]
    \setlength{\abovecaptionskip}{-0.0cm} % height above Figure X caption
    \setlength{\belowcaptionskip}{-0.2cm}
    \centering
    \includegraphics[width=2.0\columnwidth]{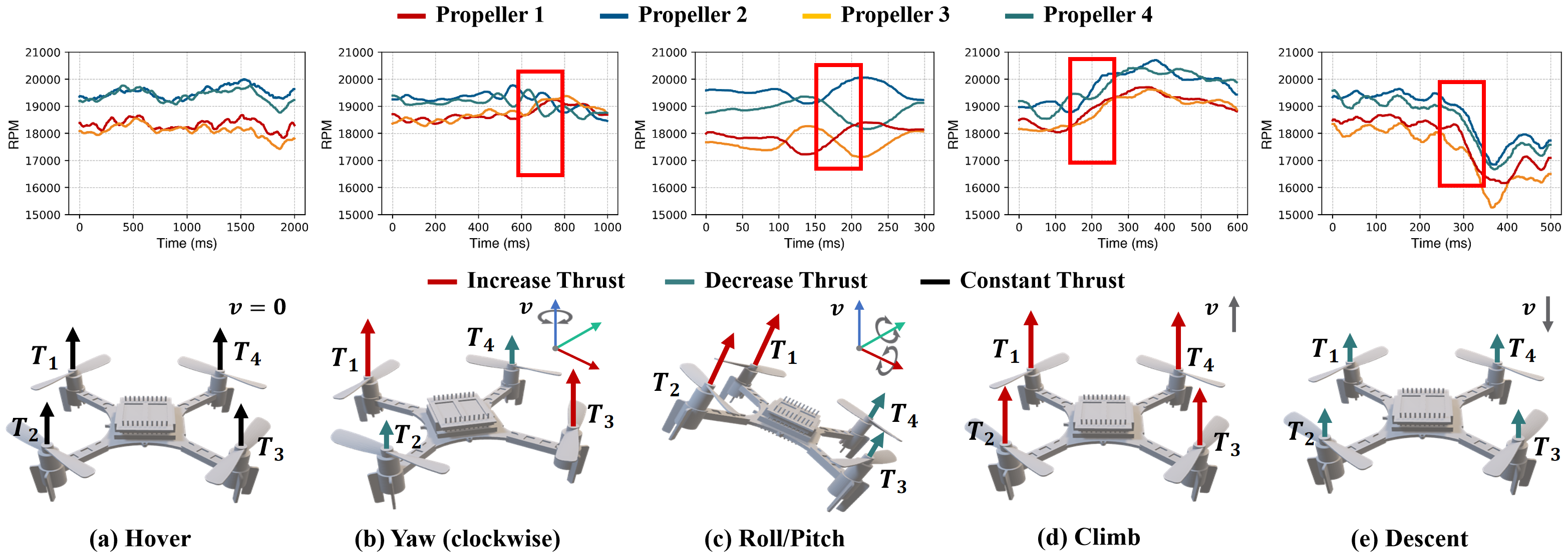}
    % \caption{Illustration of drone propeller speed distributions and the corresponding basic drone motions.}
    \caption{Illustration of basic drone motions and the corresponding propeller speeds during motion transitions.}
    \label{fig:speeddist}
    \vspace{-2mm}
\end{figure*}
\subsubsection{\textbf{Structure-aware Event Batch Downsampling}}
With the above dynamic batch size strategy, our estimate algorithm can adapt to various scenarios. However, an event batch may still contain thousands of events, causing computational strain on resource-limited devices. Inspired by existing sampling strategies for 3D point cloud processing \cite{lin2024e2pnet}, we design a structure-aware sampling approach called density importance sampling, which effectively retains spatiotemporal information, as depicted in \fig \ref{fig:batch}b. The basic idea of this method is prioritizing sampling from denser regions of the event point cloud, as these areas are less likely to contain noise and more informative for estimation.

Specifically, the following three crucial steps are involved:\\
\noindent $\bullet$ Define Neighborhood: Establish the radius of the spatiotemporal window, and identify all points $\mathcal{N}_k(e_l)$ within this radius from a given point $e_l$. \\
\noindent $\bullet$ Compute Local Density: For each point $e_l$ in the event batch, calculate its local density $\phi(e_l)$ by summing its neighbors, which can be approximated as $\phi(e_l)=\left\|\mathcal{N}_k(e_l)\right\|$.\\
\noindent $\bullet$ Density-based Sampling: Sample the event cloud based on the local density of each point. Regions with higher density will have a higher probability of being sampled. 

% \vspace{-0.2cm}
\section{Every Rotation Counts}\label{sec:erc}
\begin{figure}
    \setlength{\abovecaptionskip}{-0.0cm} % height above Figure X caption
    \centering
    \includegraphics[width=0.95\columnwidth]{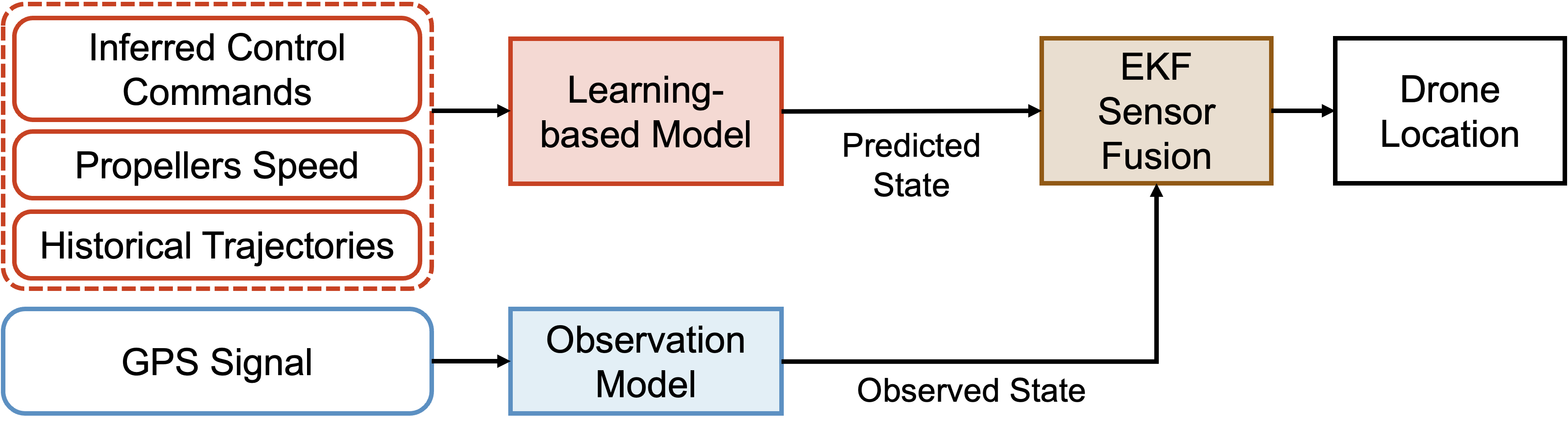}
    % \caption{The magnitude-aware reward for different candidate parameters $\omega_t$: (a) a bit smaller than ground truth, (b) ground truth, (c) a bit larger than ground truth.}
    \caption{Multi-modal fusion for drone localization.}
    \label{fig:multi-modal}
    \vspace{-6mm}
\end{figure}
The propeller rotational speeds estimated by \sysname serve as more than mere physical measurements; they encode informative representations of a drone’s internal control states, particularly relating to thrust and torque generation.
This section demonstrates how these accurate rotational signals can be leveraged to achieve two complementary objectives: ($i$) inferring drone control commands, such as pitch, roll, yaw (\S \ref{sec:intention}); and ($ii$) enhancing drone localization accuracy through dynamics-informed sensor fusion (\S \ref{sec:localization}). Rather than treating these as isolated applications, we design \textit{Every Rotation Counts} as a downstream analysis module of \sysname, driven by the rotational estimates from \textit{Count Every Rotation}.

\subsection{Drone Internal Command Inference}\label{sec:intention}

We demonstrate the \sysname' ability to infer the internal flight commands of drones based on the drone's system-level dynamics.
Existing drone monitoring systems primarily focus on detecting or localizing the drone \cite{abedi2022non, chi2022wi, karanam2018magnitude}, but lack reliable methods for estimating their intention. 

\begin{figure*}[!t]
    \setlength{\abovecaptionskip}{0.0cm} % height above Figure X caption
    \setlength{\belowcaptionskip}{-0.2cm}
    \centering
        \subfigure[Indoor experiment setup]{
            \centering
            \includegraphics[width=0.93\columnwidth]{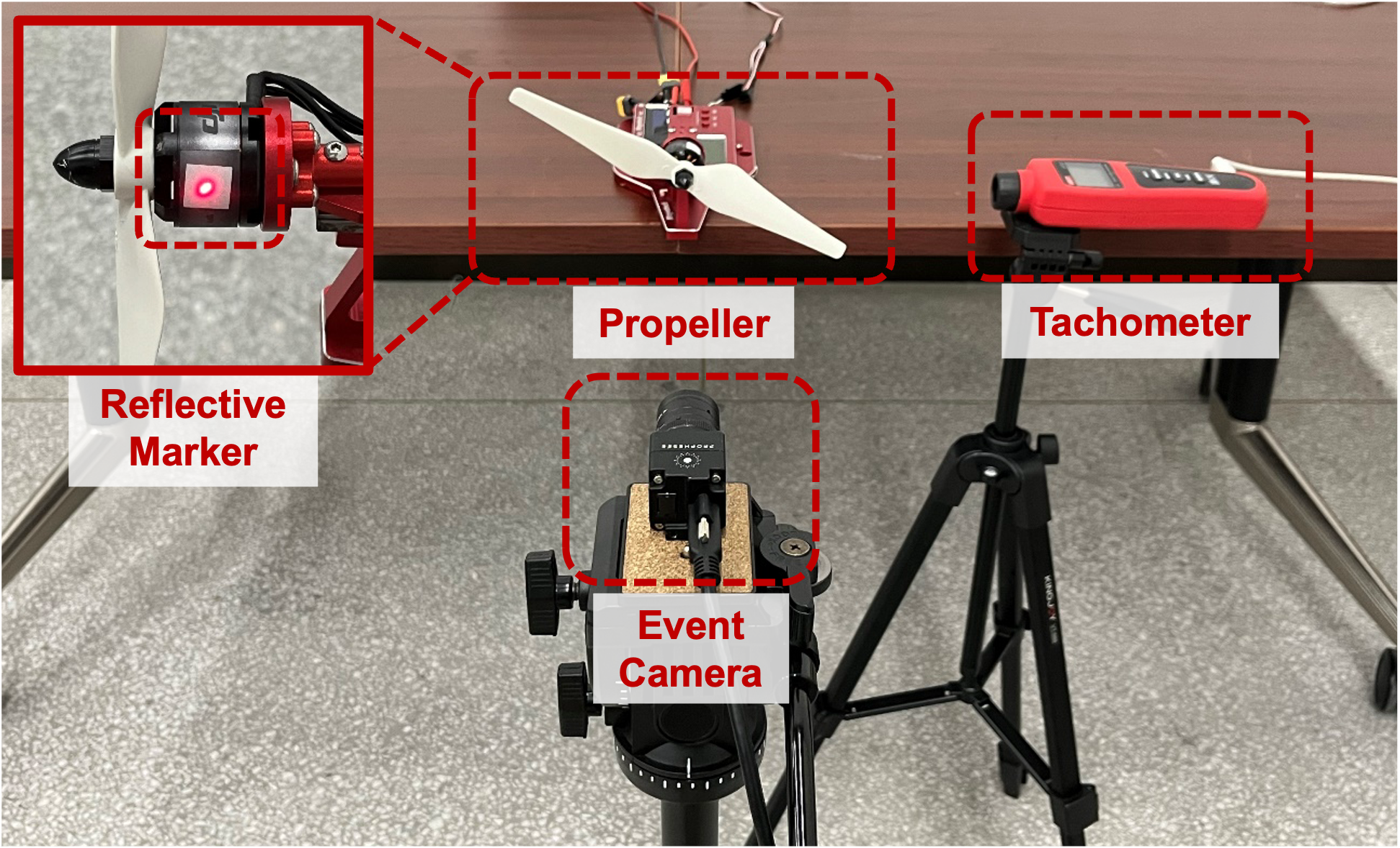}
            %\caption{fig1}
            % \label{fig:indoor_acc}
        }%
        \hspace{1mm} 
        \subfigure[Outdoor experiment setup]{
            \centering
            \includegraphics[width=0.92\columnwidth]{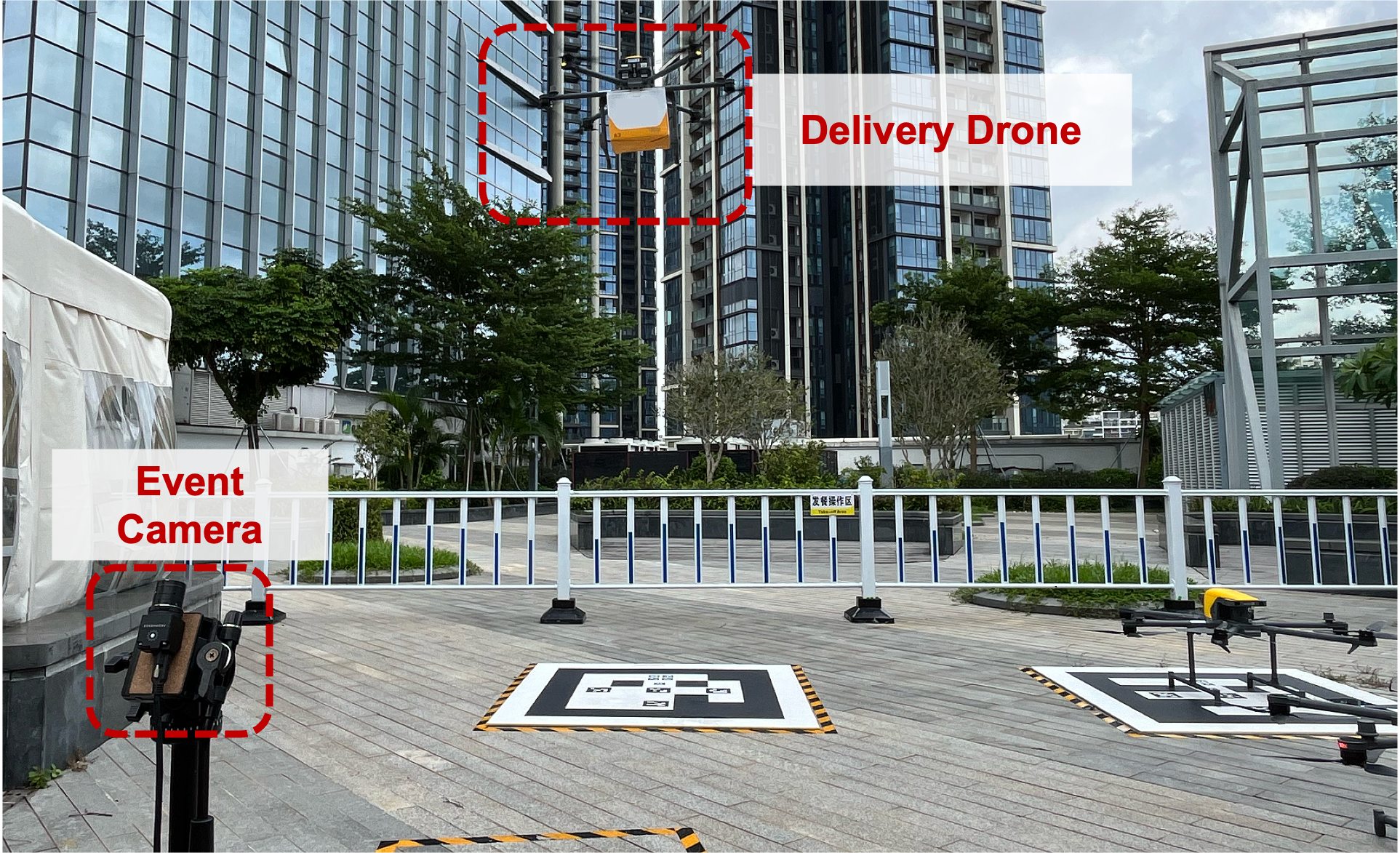}
            %\caption{fig1}
            % \label{fig:indoor_acc}
        }%
    \caption{Experimental scenarios of \sysname.}
    \label{fig:setup}
    \vspace{-2mm}
\end{figure*}
\noindent \textbf{Multirotor motion dynamics primer}. 
We elucidate the motion dynamics by considering the most common drone structure, \ie, quadrotor. Each propeller $i$ rotates with a velocity $\omega_i$ relative to the vehicle body, generating a thrust force $T_i=\kappa_{f, i} \omega_i^2$ and a torque $\tau_i=\kappa_{\tau, i} \omega_i^2$, where $\kappa_{f, i}, \kappa_{\tau, i}$ are constant parameters. Assuming equal propeller radii, the linear acceleration $a$ and angular acceleration $\alpha$ scale as:
\vspace{-0.5em}
\begin{equation}\label{eq:dynamics}
\begin{aligned}
a &\sim \frac{\sum_i T_i}{m} =\frac{\sum_i\omega_i^2d^4}{d^3}=\sum_i\omega_i^2d,\\
\alpha &\sim \frac{\sum_i \tau_i}{J}=\frac{\sum_i\omega_i^2d^5}{d^5}=\sum_i\omega_i^2,
\end{aligned}
\end{equation}
where $d$ is the characteristic length, $m$ is the drone mass, and $J$ is the drone’s mass moment of inertia \cite{mahony2012multirotor}.

\noindent \textbf{Dynamics-informed input}.
Drone flight commands can be abstracted into six basic motions: hover, yaw, roll, pitch, climb, and descent \cite{sun2022aim}. To demonstrate the feasibility of flight command inference, we conducted experiments with a quadcopter \cite{crazyfile} that performs various motions for more intuitive analysis. 
\fig \ref{fig:speeddist} depicts the variation in propeller speed distribution corresponding to changes in drone motion, revealing two key observations: ($i$) during the transition of the drone motion state, there is a noticeable change in propeller speed distribution; and ($ii$) a clear correlation exists between drone motion and its propeller speed distribution. 
These findings suggest that monitoring the propeller speed over a short period can be a feasible solution for inferring drone flight commands. Consider a drone with $N_p$ propellers, we construct the input representation $x_t = [\omega_{1, t-T_h:t}^2, \omega_{2, t-T_h:t}^2, \cdots, \omega_{N_p, t-T_h:t}^2]$, where $t$ represents the current time step, and $T_h$ is the window length that determines the amount of historical data included. 
% Thus, $x_t$ constitutes a temporal series signal with $N_p$ dimensions, enabling the prediction of future intentions.

\noindent \textbf{Lightweight prediction}.
To develop a lightweight method for predicting drone flight commands without requiring complicated structural or thrust coefficient identification, we employ a support vector machine (SVM) as an efficient classifier. Unlike neural networks that demand extensive training data and computational resources, the SVM leverages dynamics-informed inputs to establish an optimal decision boundary with superior sample efficiency and interpretability, enabling real-time flight intention prediction.
The pipeline comprises three key steps:
(\textit{i}) Filtering: apply low-pass filtering to each rotational speed signals to eliminate high-frequency noise;
(\textit{ii}) Feature extraction: extract five standardized spatial and frequency domain features from the raw speed signals: mean speed, standard deviation, dominant frequency, total spectral energy, and spectral entropy. 
(\textit{iii}) 
Training \& validation: 
Train the SVM using the extracted features and evaluate performance via k-fold cross-validation.

\subsection{Drone External Status Estimation}\label{sec:localization}
We demonstrate the feasibility of using \sysname to enhance drone localization. Multirotor autopilots typically rely on onboard sensors or GPS for localization. However, onboard sensors can quickly accumulate errors, and GPS performance degrades in urban canyons at lower altitudes, leading to instability~\cite{wang2022micnest, wang2025aerial, wang2024transformloc}. 
Our insight is that the real velocities of a drone's propellers are fundamental in determining its kinematic state. By integrating the inferred control commands from propeller speed, we can achieve high-precision, low-latency localization that directly enables precise landing assistance. The estimated drone position provides real-time corrective guidance during descent, allowing the drone to adjust its trajectory and accurately align with the target landing pad.

Our design goal is to achieve high-precision, real-time drone localization. Given the widespread adoption of GPS-based positioning for outdoor applications, we propose a cooperative approach wherein GPS, when available, can augment event camera measurements to enhance localization reliability and accuracy. Specifically, we first predict the drone's position based on its inferred control commands and propeller speed, then fuse the estimated position with the measurement of GPS via Extended Kalman Filter (EKF).

\noindent \textbf{Learning-based position prediction model}.
We develop a deep-learning model to predict a drone's short-term trajectory, leveraging inferred control command and temporal motion patterns. For a multirotor with $N_p$ propellers flying in the scene, the input state to the model consists of three parts. 
The first part is the predicted control command $u^t$ at current time $t$, represented as a one-hot encoded vector to capture the drone’s future intent.
The second part comprises the rotational speeds of the drone's propellers, denoted as $\omega^{t}_{1:N_p}$.
The third part is the drone's previous trajectory over the past $t$ time steps, represented as a sequence of coordinates: $X^{1: t}=\left\{\left(x^\tau, y^\tau, z^\tau \right) \mid \tau=1, \cdots, t\right\}$. This trajectory sequence encodes temporal velocity and acceleration patterns inherent to the drone’s motion. 
Our aim is to predict the drone's future positions for the next $T$ time steps, \ie, $Y^{1: T}=\left\{\left(x^\tau, y^\tau, z^\tau\right) \mid \tau=t+1, \cdots, t+T\right\}$. 

We first adopt a Long Short Term Memory (LSTM) network to encode the temporal dependency of historical information into a high dimensional feature for time $t$. Next, we construct another LSTM as the decoder to predict the drone's future positions. The model is trained using L2 loss, which measures the distance between the ground truth position $\hat{Y^\tau}$ and the predictive position $Y^{\tau}$. 
By jointly learning control semantics and kinematic evolution, the model bridges intent-driven actions with physical dynamics, ensuring robust short-horizon forecasts under diverse flight regimes.

\noindent \textbf{Multi-modal fusion}.
Next, we utilize the EKF~\cite{thrun2002probabilistic} algorithm to integrate inputs from two sources: GPS, and the predictive positions. EFK is widely regarded as a robust solution in nonlinear systems. It models the drone's position as a belief $bel(Y^t)$ with the mean $\mu_t$ and the covariance $\Sigma_t$.
To update these parameters, EFK consisted of two primary steps, including a \textit{prediction} step and an \textit{update} step. 
As shown in \fig \ref{fig:multi-modal}, in the \textit{prediction} step, we first utilize the learning-based model to obtain the prediction result $\mu_t^p$. In the \textit{update} step, the GPS measurements are used as observations to update the belief. 
This two-step process is repeated once there are new input data received \cite{thrun2002probabilistic}, and the mean $\mu_t^o$ of the updated belief is taken as the current estimated position.

% \vspace{-0.1cm}
\begin{figure*}[t]
\setlength{\abovecaptionskip}{-0.1cm} % height above Figure X caption
\setlength{\belowcaptionskip}{-0.4cm}
\setlength{\subfigcapskip}{-0.3cm}
    % \begin{minipage}[b]{0.9\columnwidth}
    \centering
        \subfigure[\textbf{Indoor Accuracy}]{
            \centering
            \includegraphics[width=0.5\columnwidth]{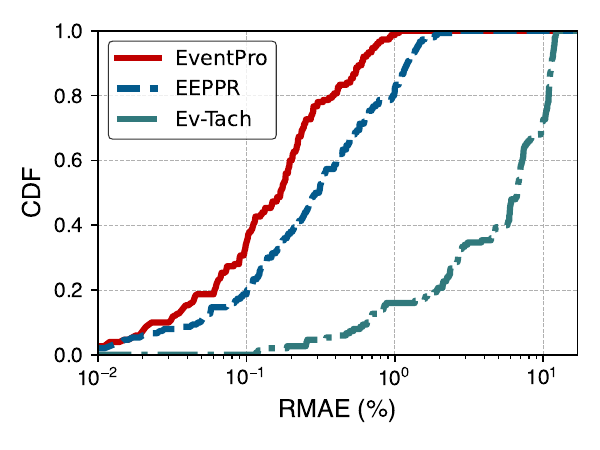}
            %\caption{fig1}
            \label{fig:indoor_acc}
        }%
        \subfigure[\textbf{Indoor Latency}]{
            \centering
            \includegraphics[width=0.5\columnwidth]{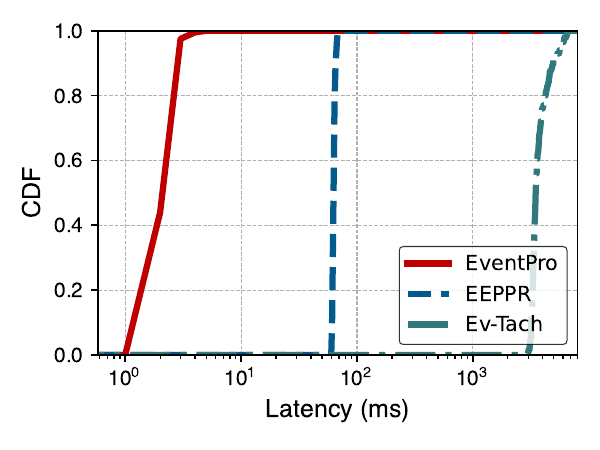}
            \label{fig:indoor_lat}
        }%
        \subfigure[\textbf{Outdoor Accuracy}]{
            \centering
            \includegraphics[width=0.5\columnwidth]{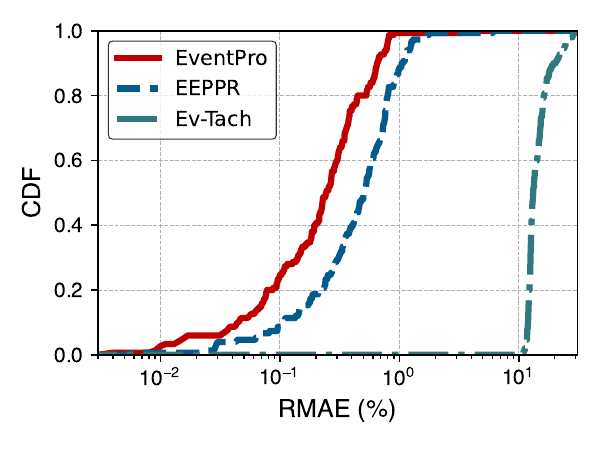}
            %\caption{fig1}
            \label{fig:outdoor_acc}
        }%
        \subfigure[\textbf{Outdoor Latency}]{
            \centering
            \includegraphics[width=0.5\columnwidth]{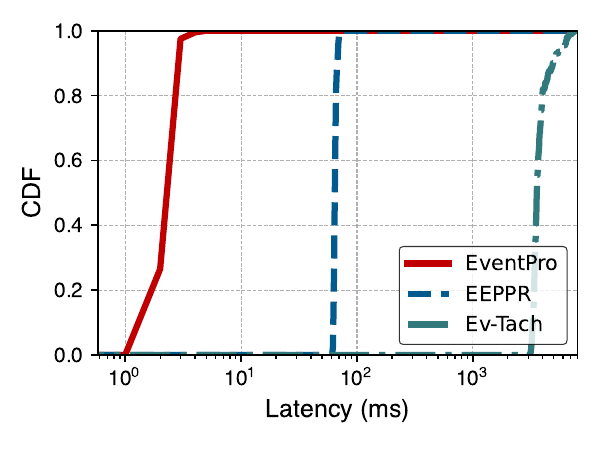}
            \label{fig:outdoor_lat}
        }%
    \caption{Overall Performance Comparison.}
    \label{fig:Overallperf}
    \vspace{-0.1cm}
\end{figure*}
\begin{figure*}
\setlength{\abovecaptionskip}{-0.1cm} % height above Figure X caption
\setlength{\belowcaptionskip}{-0.1cm}
\setlength{\subfigcapskip}{-0.3cm}
    % \begin{minipage}[b]{0.9\columnwidth}
    \centering
        \subfigure[\textbf{Impact of Rotational Speed}]{
            \centering
            \includegraphics[width=0.5\columnwidth]{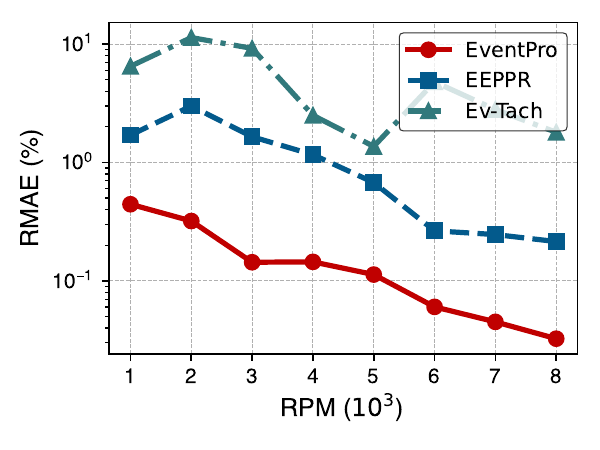}
            %\caption{fig1}
            \label{fig:impact_rpm}
        }%
        \subfigure[\textbf{Impact of Estimation Distance}]{
            \centering
            \includegraphics[width=0.5\columnwidth]{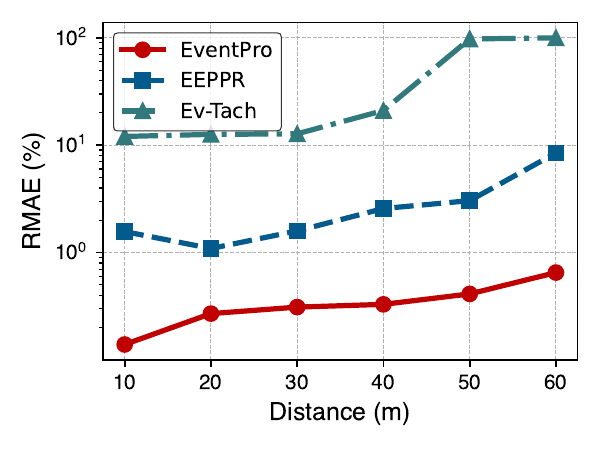}
            \label{fig:impact_dis}
        }%
        \subfigure[\textbf{Impact of View Angle}]{
            \centering
            \includegraphics[width=0.5\columnwidth]{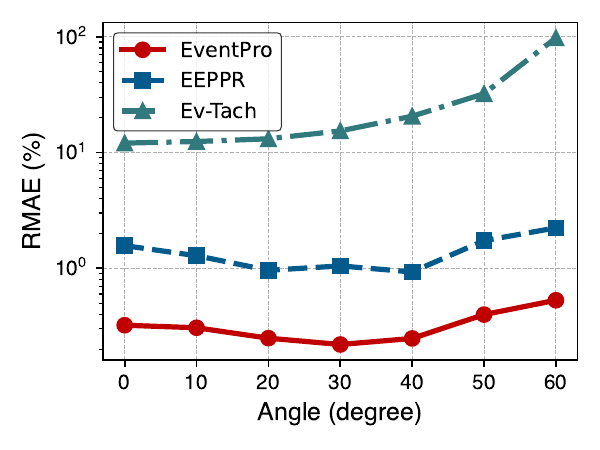}
            %\caption{fig1}
            \label{fig:impact_ang}
        }%
        \subfigure[\textbf{Impact of Illumination}]{
            \centering
            \includegraphics[width=0.5\columnwidth]{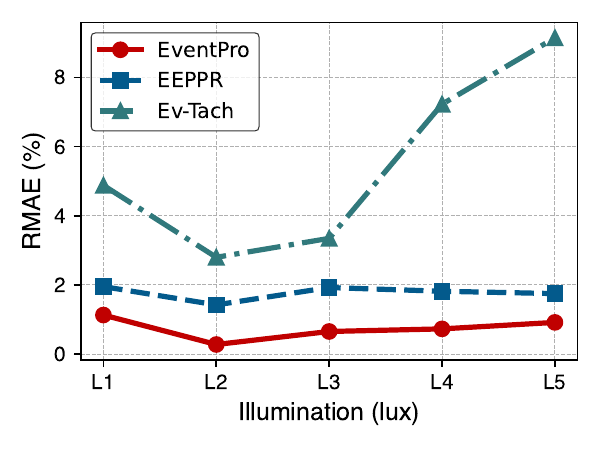}
            \label{fig:impact_noi}
        }%
    \caption{System Robustness Evaluation.}
    \label{fig:Robustness}
    \vspace{-0.0cm}
\end{figure*}
\section{Implementation and Evaluation}
\subsection{Implementation}
As shown in \fig \ref{fig:setup}, we implement \sysname with a commercial event camera, the Prophesee EVK4 HD evaluation kit \cite{evk4}, which is equipped with the IMX636ES event-based vision sensor for high-definition event data capture (1280 × 720 pixels) using the Soyo SFA0820-5M lens. Our implementation contains more than 20000 LOC (lines of Python code). \sysname operates on a laptop equipped with an Intel i7-12900K CPU, and 32GB RAM, running on the Ubuntu 20.04 operating system. The event camera is connected to the laptop via USB 3.0. 
All data sampling, streaming, and computation are performed in real-time on a laptop, enabling accurate and efficient speed estimation of the drone’s propellers.

% \vspace{-0.3cm}
\subsection{Benchmark Evaluation}
\subsubsection{Experimental Methodology}\label{sec:allsetup}
\ \\
\noindent \textbf{Experiment Setting}.
We conduct field studies both indoor and outdoors as shown in \fig \ref{fig:setup}.

\noindent $\bullet$ \textit{Indoor experiments}. We build a laboratory testbed featuring a rotating propeller mounted on a brushless motor shaft. The setup comprises a laptop and an Arduino Uno to control the motor's speed by adjusting the duty cycle and frequency of the PWM signal. The laptop transmits the PWM parameters to the Arduino Uno via a serial port, which then relays the PWM signal to the motor's Electronic Speed Controller (ESC). An event camera was positioned in front of the propeller. 

\noindent $\bullet$ \textit{Outdoor experiments}. We adopt a commercial delivery drone and conduct real-world experiments in a test field. This drone is manufactured by a company specializing in daily and city-wide instant delivery services. The drone features six propellers attached to brushless motors and is controlled by a PX4\cite{px4} flight controller. An event camera was positioned on a ground-based bracket to capture the drone in flight.
We conducted extensive experiments over 10 hours, collecting more than 600GB of raw event data.

\begin{figure*}[t]
\setlength{\abovecaptionskip}{-0.1cm} % height above Figure X caption
\setlength{\subfigcapskip}{-0.3cm}
    % \begin{minipage}[b]{0.9\columnwidth}
    \centering
        \subfigure[\textbf{Prediction Accuracy}]{
            \centering
            \includegraphics[width=0.5\columnwidth]{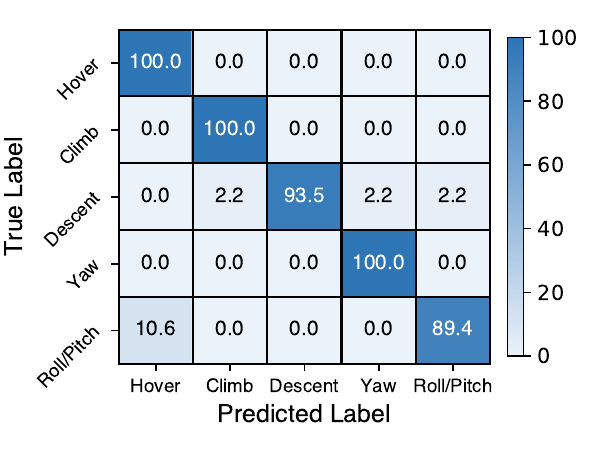}
            %\caption{fig1}
            \label{fig:intention:svm}
        }%
        \subfigure[\textbf{Prediction Latency}]{
            \centering
            \includegraphics[width=0.5\columnwidth]
            {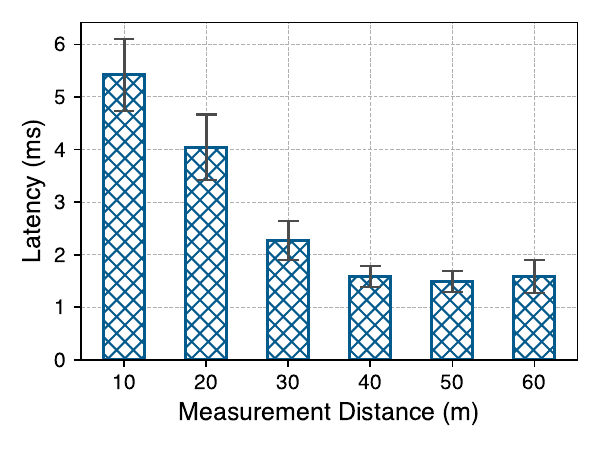}
            \label{fig:intention:latency}
        }%
        \subfigure[\textbf{Impact of Motion Speed}]{
            \centering
            \includegraphics[width=0.5\columnwidth]{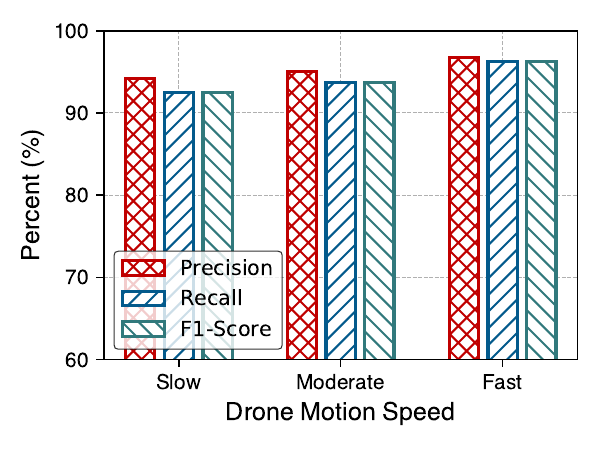}
            \label{fig:intention:impact_v}
        }%
        \subfigure[\textbf{Impact of Classification Model}]{
            \centering
            \includegraphics[width=0.5\columnwidth]{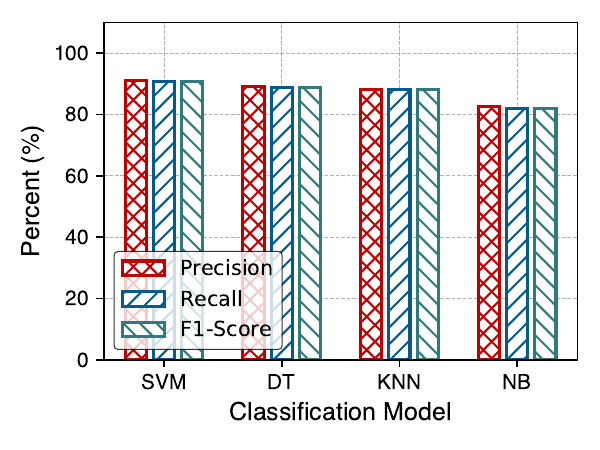}
            %\caption{fig1}
            \label{fig:intention:impact_mod}
        }%
    \caption{Experiment results of drone internal command inference.}
    \label{fig:intention}
    % \vspace{-0.2cm}
\end{figure*}
\noindent \textbf{Metrics and Ground Truth}.
Similar to related works, we use the relative mean absolute error (RMAE) to evaluate the estimation accuracy: $RMAE = \frac{1}{N}\sum_{i=1}^N \frac{|r_i-r_{gt}|}{r_{gt}} \times 100\%$, 
where $N$ is the number of experiments and N=100, $r_i$ is the $i_{th}$ estimated rotational speed and $r_gt$ is the ground truth. Lower RMAE indicates higher measurement accuracy.

The ground truth for indoor experiments is obtained using a high-precision laser tachometer (UNI-T UT372\cite{UT372}) (\fig \ref{fig:setup}a) with a relative error of ±0.04\%. A reflective marker is affixed to the motor shell to reflect infrared light, and the tachometer is mounted on a tripod aimed at the marker. For outdoor experiments, a built-in magnetic encoder attached to the shaft of the drone's motor provides ground truth with a relative error of ±0.5\%. All measurement are synchronized with \sysname and baselines for comparison.

\noindent \textbf{Baselines}.
We compare \sysname with two approaches: 1) \textbf{Ev-Tach} \cite{zhao2023ev}, a state-of-the-art system that leverages the classicial iterative closest point (ICP) algorithm for rotational speed estimation; 
2) \textbf{EEPPR} \cite{kolavr2024ee3p3d}, a cutting-edge method that measures the period of rotational motion by analyzing the 3D correlation of events within a spatiotemporal window.

\subsubsection{Overall Performance}
\ \\
\noindent \textbf{Estimation accuracy}.
\fig \ref{fig:Overallperf}a illustrates the estimation accuracy of \sysname compared to two baselines in an indoor testbed. \sysname achieves an average RMAE of 0.23\%, surpassing both Ev-Tach and EEPPR, which have RMAEs of 0.48\% and 6.19\%, respectively. The performance of these systems is also compared in an outdoor scenario, as shown in \fig \ref{fig:Overallperf}c. \sysname records the lowest average error of 0.31\%, outperforming the two baselines by 45.9\% and 97.9\%, respectively. 
Unlike Ev-Tach, which relies on point-to-point matching that is prone to noise interference, \sysname uses an event set-based method for data association, enhancing its estimation accuracy. Additionally, \sysname employs a carefully designed objective function that accounts for propeller geometry, resulting in improved noise resistance.

\noindent \textbf{End-to-end latency}.
We further assess the end-to-end latency in both indoors and outdoors scenarios. As shown in \fig \ref{fig:Overallperf}b, \sysname achieves an average latency of 3.14$ms$, outperforming both baselines (63.90$ms$ and 3.8$s$, respectively). \fig \ref{fig:Overallperf}d illustrates the outdoor latency comparison, where \sysname again leads with the lowest average latency of 3.22$ms$, outperforming EEPPR by one order of magnitude and EV-Tach by three orders of magnitude.
The baselines experience an inherent increase in latency due to the burst of events, especially at high rotational speeds. The large volume of event data necessitates a costly registration step for event alignment, which delays the systems. In contrast, \sysname dynamically extracts a lightweight representation from event streams while preserving the crucial space-time structure, thereby reducing the latency without compromising estimation accuracy.

\renewcommand{\arraystretch}{1.0}  % 表格行距的调整，可选
\begin{table}[]
\setlength{\abovecaptionskip}{-0.00cm} % height above Figure X caption
\setlength{\belowcaptionskip}{-0.5cm}
% \small
\caption{Different Drone Motion Speed Configurations}
\label{tab:tab1}
\centering
% \resizebox{\columnwidth}{!}{%
\fontsize{8}{8}\selectfont
\begin{tabular}{c@{\hspace{0.8cm}}cc}  % 调整@{\hspace{1.5cm}}中的距离值按需要变化
\toprule
\multirow{2}{*}{Motion Mode} & \multicolumn{2}{c}{Speed}          \\  \cmidrule(lr){2-3} 
                             & Translation (m/s) & Rotation (°/s) \\ \cmidrule(lr){1-3} 
Slow                         & 0-1.0             & 0-20.0         \\
Moderate                     & 3.0-5.0           & 30.0-60.0      \\
Fast                         & 8.0-10.0          & 70.0-100.0     \\ \bottomrule
\end{tabular}
% }%
\vspace{-0.35cm}
\end{table}

\subsubsection{System Robustness Evaluation}\label{sec:robustness}
\noindent \textbf{Impact of rotational speed}.
\fig \ref{fig:Robustness}a depicts the impact of rotational speed within the typical operating range of drone \cite{deters2017static}. \sysname consistently outperforms baselines across all tested settings. As rotational speed increases, we observe a slight downward trend across all methods. This is due to slower speeds amplifying the impact of the same error, leading to higher RMAE. In contrast, \sysname consistently maintains stable and satisfactory estimation performance regardless of rotational speed changes.

\noindent\textbf{Impact of measurement distance}.
We further verify \sysname's effectiveness at various distances under the outdoor experiment setup. As illustrated in \fig \ref{fig:Robustness}b, the RMAE of all methods increases with distance, yet \sysname consistently outperforms both baselines. At the adverse condition where the measurement distance reaches $60m$, the propeller occupies less than 10\% of the event camera's field of view, causing propeller-triggered events to be obscured by noise. Consequently, \sysname's RMAE increases to 0.65\%, though performance remains acceptable. In contrast, Ev-Tach's performance deteriorates significantly beyond $30m$ due to insufficient events for point registration. These results demonstrate \sysname's robustness for propeller speed estimation under diverse environmental conditions.

\noindent\textbf{Impact of view angle}.
The view angle critically impacts the number of triggered events per propeller rotation, thereby influencing the available features for speed estimation. We conduct experiments across various view angles to assess their effect on estimation accuracy, as shown in Fig.~\ref{fig:Robustness}c. Larger view angles cause a substantial increase in RMAE for all methods, particularly when exceeding 40 degrees. Nevertheless, even under the worst-case scenario, \sysname maintains superior accuracy compared to baseline approaches, demonstrating its robustness against event feature degradation. Technically, as long as individual propeller blades are clearly distinguishable in the event camera's field of view, our proposed dynamic batch size strategy can resolve data association issues and achieve accurate speed estimation.

\noindent\textbf{Impact of illumination}.
As a vision-based technique, robustness under varying illumination conditions is critical. Based on extensive field measurements, we established five representative lighting levels for our experimental setup: L1 (10 lux), L2 (100 lux), L3 (1000 lux), L4 (10,000 lux), and L5 (100,000 lux), covering scenarios ranging from dusk and cloudy days to extreme sunny midday conditions. Results depicted in Fig. \ref{fig:Robustness}d present the RMAE performance. All methods experience increased RMAE under extremely weak illumination. This degradation is expected because event cameras are highly sensitive to light changes, and these boundary conditions generate more noise, thereby negatively impacting speed estimation performance. Notably, \sysname exhibited a significantly lower RMAE of 1.12\% compared to 1.95\% and 4.88\% for EEPPR and Ev-Tach, even under the most challenging L1 illumination scenario. This performance demonstrates \sysname's superiority across various operational illumination conditions.

\subsection{Drone Internal Command Inference Evaluation}
\subsubsection{Experimental Methodology}
We conduct outdoor experiments as described in \S \ref{sec:allsetup} and \fig \ref{fig:setup}b. 
For each drone motion, 200 samples lasting about 100 microseconds each were collected for training.
We employed accuracy and latency as the primary metrics to evaluate performance. 
\subsubsection{Experiment Results}
\quad\\
\noindent \textbf{Accuracy and latency}.
\fig \ref{fig:intention}a depicts the confusion matrix of accuracy for various drone motions, demonstrating high accuracies exceeding 89\% across all scenarios. Notably, the roll/pitch motion includes more complex drone angular acceleration changes, complicating classification and leading to occasional misidentifications.
\fig \ref{fig:intention}b illustrates the prediction latency across different measurement distances, consistently below 5.5$ms$ in all scenarios. The results indicate that latency decreases as distance increases, attributable to the reduced number of events requiring processing. 

\begin{figure*}[t]
\setlength{\abovecaptionskip}{-0.1cm} % height above Figure X caption
\setlength{\belowcaptionskip}{-0.2cm}
\setlength{\subfigcapskip}{-0.3cm}
    % \begin{minipage}[b]{0.9\columnwidth}
    \centering
        \subfigure[{\textbf{Performance Comparison}}]{
            \centering
            \includegraphics[width=0.66\columnwidth]{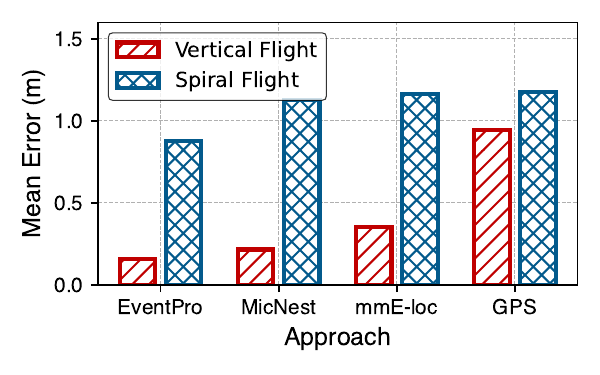}
            %\caption{fig1}
            \label{fig:loc:overall}
        }%
        \subfigure[\textbf{Impact of Altitude}]{
            \centering
            \includegraphics[width=0.66\columnwidth]{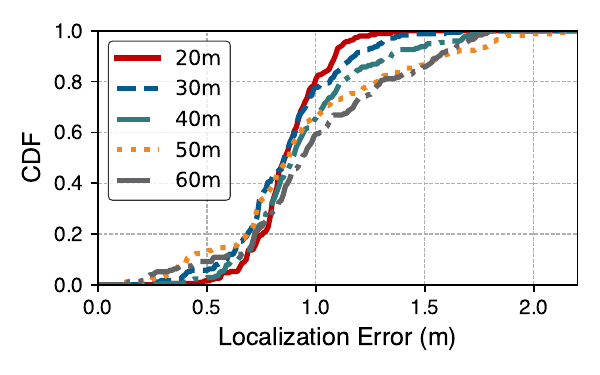}
            %\caption{fig1}
            \label{fig:loc:impact_alt}
        }%
        \subfigure[\textbf{Impact of Drone Acceleration}]{
            \centering
            \includegraphics[width=0.66\columnwidth]{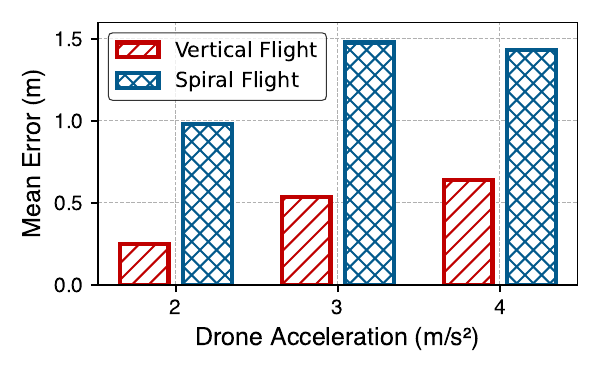}
            \label{fig:loc:impact_acc}
        }%
        % \subfigure[\textbf{Impact of Wind}]{
        %     \centering
        %     \includegraphics[width=0.5\columnwidth]{evalFigs/localization/mp_diff_wind.pdf}
        %     %\caption{fig1}
        %     \label{fig:loc:impact_wind}
        % }%
    \caption{Experiment results of drone external status estimation.}
    \label{fig:localization}
    \vspace{-0.2cm}
\end{figure*}

\begin{figure}
\setlength{\abovecaptionskip}{-0.1cm} % height above Figure X caption
\setlength{\belowcaptionskip}{-0.2cm}
\setlength{\subfigcapskip}{-0.3cm}
    % \begin{minipage}[b]{0.9\columnwidth}
    \centering
        % \subfigure[\textbf{Impact of Drone Acceleration}]{
        %     \centering
        %     \includegraphics[width=0.50\columnwidth]{evalFigs/localization/mp_diff_acc.pdf}
        %     \label{fig:loc:impact_acc}
        % }%
        \subfigure[\textbf{Impact of Wind}]{
            \centering
            \includegraphics[width=0.5\columnwidth]{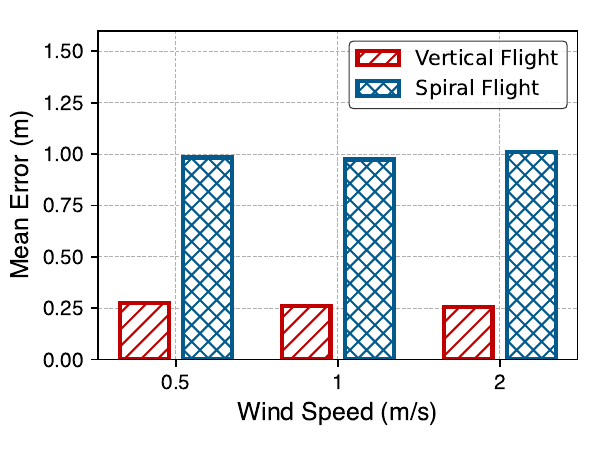}
            %\caption{fig1}
            \label{fig:loc:impact_wind}
        }%
        \subfigure[{\textbf{Impact of Background Motion}}]{
            \centering
            \includegraphics[width=0.5\columnwidth]{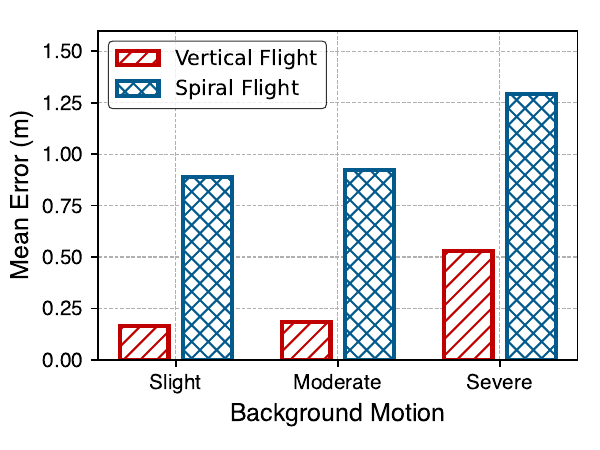}
            %\caption{fig1}
            \label{fig:loc:impact_bgmotion}
        }%
    \caption{Drone external status estimation performance under dynamic environmental conditions.}
    \label{fig:localization_robustness}
    \vspace{-0.8cm}
\end{figure}

\noindent \textbf{Impact of drone motion speed}.
We evaluate the system's effectiveness across various drone motion speed, as defined in \tab \ref{tab:tab1}. The results shown in \fig \ref{fig:intention}c indicate that accuracy increases with higher motion speed. This outcome aligns with expectations, as greater motion speeds produce more distinguishable patterns in the propeller speed distribution, thereby facilitating flight command inference.

\noindent \textbf{Impact of classification model}.
We investigate the impact of various classification models on system performance by implementing three additional classifiers: decision tree (DT), k-nearest neighbors (KNN), and naive Bayes (NB). As shown in \fig \ref{fig:intention}d, all models maintain satisfactory performance despite accuracy variations. We anticipate that more advanced techniques, such as deep neural networks (DNN), could further enhance prediction performance.

\subsection{Drone External Status Estimation Evaluation}
\subsubsection{Experimental Methodology}
The experiment setup is depicted in \fig \ref{fig:setup}b. A Real-Time Kinematic Positioning (RTK) base station was deployed to provide high-fidelity localization results as the ground truth. Following standard practice in precision landing systems~\cite{wang2025ultra}, we adopt mean localization error in meters as the primary metric, defined as the average positioning error across latitude, longitude, and altitude. This metric directly reflects landing performance: larger errors may result in landing pad misalignment, failed connection with charging ports, or necessitate abort maneuvers.
We compare the mean error of \sysname against three baselines: ($i$) GPS; ($ii$) MicNest~\cite{wang2022micnest}, a state-of-the-art acoustic localization method; and ($iii$) mmE-loc~\cite{wang2025ultra} a state-of-the-art localization system based on event cameras and mmWave radar fusion.

\noindent \textbf{Flight trajectories}. The drone operates automatically according to pre-planned trajectories from the ground station. We consider two classical flight trajectories following previous work~\cite{wang2022micnest}:
1) \textit{Vertical flight}, where the drone takes off, climbs vertically to a specified altitude, and then lands back onto the platform;
% 2) \textit{Horizontal flight}, where the drone follows a squared spiral trajectory at a constant altitude.
2) \textit{Spiral flight}, where the drone ascends to a fixed altitude, performs three loops along a complex squared spiral path, and returns to land on the platform.
The default flight speed is 4$m/s$.
% \vspace{-0.35cm}
\subsubsection{Experimental Results}
\quad\\
\textbf{Localization performance}.
\fig \ref{fig:localization}a presents the localization performance of \sysname compared to baseline methods. The specified altitude for both flight types in the experiments is 50m. We conducted repeated experiments under diverse weather conditions (\eg, cloudy and sunny) with illumination from 1000 to 100,000 lux and computed the average results. \sysname achieved an average localization error of 0.15$m$ for vertical flights and 0.87$m$ for spiral flights, outperforming all baselines. The error was lower during vertical flights, attributed to more consistent motion intent and closer proximity to the ground platform compared to spiral flights.

\noindent \textbf{Impact of altitude}.
\fig \ref{fig:localization}b illustrates the cumulative distribution function (CDF) of the localization error across various flight altitudes. The results indicate that \sysname successfully localizes the drone, with an average localization error of only 0.93m across all altitudes. Moreover, it was observed that localization error grows as altitude increases, due to factors described in \sec \ref{sec:robustness} concerning performance at various measurement distances.

\noindent\textbf{Impact of drone acceleration}.
We study how drone acceleration affects localization performance, as propeller speeds directly influence acceleration and positioning. As shown in \fig \ref{fig:localization}c, localization error increases slightly with acceleration in both vertical and spiral flights. Specifically, mean error reaches 0.63$m$ for vertical flight at $4m/s^2$ acceleration, degrading \sysname's localization performance. This occurs because rapid dynamics amplify noise and biases in inertial sensors and introduce GPS delays and inaccuracies, complicating sensor fusion algorithms. However, this can be mitigated by appropriately tuning the EKF parameters to adjust the weight of the GPS observation model based on its actual accuracy.

\noindent\textbf{Impact of wind}.
We further evaluate the effectiveness of our system in environments with varying wind speed conditions. Generally, a higher wind speeds can cause the motors' actual angular speed to deviate from the commanded speed, leading to an unstable trajectory and complicating the localization process. However, as shown in \fig \ref{fig:localization_robustness}a, even with a wind speed of 2$m/s$, the average localization accuracy remains consistent. These results demonstrate that our system can effectively infer the real movement intention of the drone based on the actual propeller speed, making it making it a promising solution for reliable drone localization under diverse dynamic wind conditions.

% \vspace{-2mm}
\noindent\textbf{Impact of background motion}.
We examine the impact of background motion on localization performance. In real-world scenarios, various weather conditions induce dynamic background variations that similarly impact event-based propeller perception, all manifesting as dynamic background noise of varying intensities. We therefore conduct experiments under different rainfall intensities as representative background motion levels: slight ($\sim$5 $mm/h$), moderate ($\sim$10 $mm/h$), and severe ($\sim$40 $mm/h$). As shown in \fig\ref{fig:localization_robustness}b, the mean localization error remains consistent even under moderate background motion, as the event camera captures sufficient propeller-triggered events. In the most challenging scenario, mean error increases due to background noise occupying a considerable proportion of total events. Nevertheless, \sysname maintains an error of less than 1.3$m$, validating that our geometry-instructed event motion compensation algorithm enables accurate estimation robust to noise interference.

\vspace{-0.4cm}
\section{Related Work}
\noindent \textbf{Rotational speed estimation}.
Rotational motion sensing methods are broadly categorized as invasive or non-invasive.
Invasive methods, such as mechanical tachometers \cite{liptak2003instrument}, electrostatic sensors \cite{li2019digital}, Hall-effect sensors \cite{kathirvelan2014hall}, and optical encoders \cite{asfia2011high}, involve physical contact or additional hardware, potentially disrupting the system's natural state and limiting flexibility. Non-invasive methods utilize acoustic \cite{li2018pcias}, RF signals \cite{heggo2022rftacho}, WiFi \cite{lin2022wi}, optical \cite{wang2017rotational,wang2018rotational,kim2016visual,natili2020video,lasertacho}, and radar \cite{ciattaglia2023uav,ciattaglia2023determination,ding2023rotation} techniques, but they face significant limitations in spatial sensing range or temporal resolution, often struggling with noise interference. For example, laser tachometers \cite{lasertacho,cheng2011contactless} require retroreflective markers and precise aiming, 
with an operation distance under 20 cm. 
RGB camera-based methods \cite{wang2017rotational,wang2018rotational} are restricted by a 60 fps frame rate, capping measurable speeds at 1500 RPM.
Additionally, most of the above methods are designed for scenarios involving a single rotating object, which is inadequate for our settings.

The most relevant work, Ev-Tach \cite{zhao2023ev}, introduces a notable rotational speed estimation solution using event cameras. It employs a 3D points-based registration method to estimate the transformation between two slices of event streams, thereby determining the angle of rotation. However, Ev-Tach primarily targets fixed rotating objects and requires a small-scale work environment with a maximum sensing distance of around 70 cm. In contrast, our \sysname focuses on propeller speed sensing for high-speed multirotors in real-world environments. This shift from fixed targets at close distances to dynamic targets in diverse conditions, combined with outdoor noise, introduces new challenges to fully harness event cameras' potential, as elaborated in $\S$\ref{sec:intro}.

\noindent \textbf{Event-based algorithms and systems}.
Event camera-based sensing has garnered increasing popularity across various applications, owing to their high temporal resolution, low latency, and high dynamic range. Recent systems employ them for object localization~\cite{xu2023taming, wang2025event, luo2024eventtracker}, object tracking~\cite{hao2023eventboost, falanga2020dynamic, mitrokhin2018event}, SLAM~\cite{rebecq2016evo}, and motion estimation~\cite{gallego2017accurate,ng2023direct,kim2021real,shiba2024secrets,zhu2019unsupervised}. Note that though~\cite{gallego2017accurate,ng2023direct,kim2021real} focus on rotational speed estimation, they aim to measure the rotation of the event camera itself. 
Our work fundamentally differs by leveraging the event camera to estimate the rotational speed of external targets in real-world scenarios.
In \sysname, we introduce a novel motion compensation-based pipeline that incorporates propeller geometries for accurate and real-time drone propeller speed estimation, featuring three effective modules. Our design aims to improve the accuracy and efficiency of event-based systems, particularly benefiting high-speed drones.

\section{Discussion}
We discuss the limitations and future directions to broaden the applicability of \sysname.

\noindent\textbf{Deployment Conditions and Future Applications.}
\sysname is applicable to various multi-rotor platforms due to their similar rotational characteristics. However, event cameras remain in active development with certain limitations, including low resolution and fixed aperture. Based on field experiments, we identify optimal deployment conditions: observation distances under 50m, viewing angles between 0-50 degrees, and illumination below 100,000 lux. These conditions are well-suited for drone landing scenarios~\cite{zhao2025airscape, wang2025mme}. Looking forward, \sysname holds potential for broader applications, including multi-drone localization~\cite{chen2024soscheduler} and high-speed FPV trajectory prediction. To enable these applications, the system requires enhancement with clustering algorithms for distinguishing multiple drones and more robust spatiotemporal feature extraction methods for detecting small-scale drones at extended distances.

\noindent\textbf{Influence of Weather.}
Illumination is the primary environmental factor affecting event camera performance~\cite{wang2025enabling}. Extreme lighting conditions on sunny days have minimal impact on \sysname, as event cameras respond to relative rather than absolute light intensity changes. Conversely, weak illumination reduces event generation due to insufficient intensity variations, as demonstrated in \fig\ref{fig:Robustness}d. Additionally, dynamic background motion induced by fog or rain may degrade drone localization performance. A future direction is developing more robust event preprocessing algorithms that leverage spatiotemporal drone features in the frequency domain to mitigate these adverse effects~\cite{ruan2025pre, ruan2025edmamba}.

\noindent\textbf{Computational and Cost consideration.}
Our evaluation deploys \sysname on a laptop equipped with an Intel i7-12900K CPU and 32GB RAM, achieving real-time performance. Notably, modern compact edge computing platforms, such as the ASUS NUC 14 Pro~\cite{nuc}, can fully support these computational requirements, enabling flexible deployment in resource-constrained environments. Regarding cost considerations, the primary bottleneck lies in event camera hardware, with prices ranging from \$300 to \$5,000 depending on resolution, sensitivity. However, we anticipate significant price reductions in the near future, driven by rapid advancements in neuromorphic vision hardware and increasing commercial adoption. Our work presents an active and contactless solution for drone sensing, positioning our system to readily benefit from ongoing advancements in event camera technology.

\section{Conclusion}
This paper demonstrates that focusing on propeller rotational speed significantly enhances drone sensing performance. We present \sysname, an event-camera-based system with two components: \textit{Count Every Rotation}, which enables accurate, real-time propeller speed estimation by mitigating environmental noise sensitivity; and \textit{Every Rotation Counts}, which infers internal flight commands and external dynamics from estimated speeds. 
Extensive real-world evaluations in drone delivery scenarios show that \sysname achieves remarkable sensing latency and accuracy in rotational speed estimation, while demonstrating dual capabilities of inferring internal drone commands and enhancing external flight status estimation.

%%
%% The next two lines define the bibliography style to be used, and
%% the bibliography file.
\newpage
\bibliographystyle{ACM-Reference-Format}
\bibliography{ref}

%%
%% If your work has an appendix, this is the place to put it.

\end{document}